\documentclass{article}

\PassOptionsToPackage{numbers}{natbib}
\usepackage[preprint]{neurips_2026}

\usepackage[utf8]{inputenc} % allow utf-8 input
\usepackage[T1]{fontenc}    % use 8-bit T1 fonts
\usepackage{url}            % simple URL typesetting
\usepackage{booktabs}       % professional-quality tables
\usepackage{amsfonts}       % blackboard math symbols
\usepackage{amsmath}
\usepackage{amssymb}
\usepackage{nicefrac}       % compact symbols for 1/2, etc.
\usepackage{microtype}      % microtypography
\usepackage{graphicx}
\usepackage{xcolor}
\usepackage{multirow}
\usepackage{array}
\usepackage{enumitem}
\usepackage{listings}
\usepackage{tikz}
\usetikzlibrary{shapes.geometric, arrows.meta, positioning, calc, fit, backgrounds, decorations.pathreplacing}
\usepackage{float}          % [H] placement for appendix floats
\usepackage{hyperref}       % load after neurips_2026 (the sty loads natbib, not hyperref)
\usepackage{subcaption}     % after hyperref
\graphicspath{ {./images/} }

\lstdefinestyle{prompt}{
  basicstyle=\ttfamily\footnotesize,
  breaklines=true,
  breakatwhitespace=true,
  frame=single,
  framerule=0.4pt,
  rulecolor=\color{black!40},
  backgroundcolor=\color{black!2},
  xleftmargin=4pt,
  xrightmargin=4pt,
  aboveskip=6pt,
  belowskip=6pt,
  columns=fullflexible,
  keepspaces=true,
}

\newcommand{\toolname}[1]{\texttt{\small #1}}
\newcommand{\methodname}{LiteSearch-VL}
\newcommand{\dataset}{SearchVL-SFT-36k}

\title{\methodname{}: Small Multimodal Search Agents via\\Trajectory Distillation and Synthetic Step-DPO}

\author{%
  Saeed Khaki \\
  Microsoft AI \\
  \texttt{saeedkhaki@microsoft.com} \\
  \And
  Nima Safaei \\
  Ohio State University \\
  \texttt{safaei.3@osu.edu} \\
  \And
  Kamal Ginotra \\
  Microsoft AI \\
  \texttt{kamalginotra@microsoft.com} \\
}

\begin{document}
\maketitle

\begin{abstract}
Multimodal search agents answer visual questions by interleaving image understanding, web retrieval, tool use, and evidence synthesis. Strong systems exist, but in two expensive regimes: proprietary frontier models such as GPT-5 and Gemini, or large open vision-language backbones trained with substantial agentic data and reinforcement learning. We ask a different question: when released agent trajectories are distilled into much smaller backbones under a single-node budget, what is actually transferred? We study this with \methodname{}, a low-compute recipe for Qwen3-VL-2B and Qwen3-VL-4B that uses only released OpenSearch-VL trajectories, parameter-efficient LoRA adapters, and synthetic step-level preferences: DPO on GPT-5-generated hard negatives targeting five local failure modes (\emph{premature answer, wrong tool, weak query, repeated query, ignored image}). Across 12{,}400 GPT-5-judged rollouts on SimpleVQA, FVQA, LiveVQA, and VDR-Bench-testmini, the dominant effect is behavioral rather than a uniform accuracy lift: full-trajectory supervised fine-tuning transfers the \emph{agent contract}, taking the 2B model from almost never emitting a usable answer (1{,}237/1{,}240 \texttt{no\_answer} rollouts) to 28.4\% macro Pass@1, matching or slightly exceeding the off-the-shelf 4B base (25.6\%). Synthetic preference learning and compact tool distillation act as refinements rather than phase transitions (best 4B configuration: 30.8\% macro Pass@1). Finally, a controlled VDR step-budget ablation shows that extra search turns convert abstentions into \texttt{wrong\_entity} errors rather than correct answers, identifying answer verification, not search depth, as the next bottleneck for small multimodal agents.
\end{abstract}

%--------------------------------------------------------------------------
\section{Introduction}
\label{sec:intro}
Knowledge-intensive visual questions often cannot be answered from the image alone: a model may need to identify an entity in the scene, search for external evidence, inspect a crop or an OCR region, and decide when the accumulated evidence is sufficient to commit to a final answer. This has motivated multimodal search agents that interleave reasoning with retrieval and image-processing tools rather than producing a single direct response~\cite{geng2025webwatcher, huang2026visiondeepresearch, chen2026opensearchvl}. The most capable systems, however, are expensive to reproduce: they either depend on proprietary frontier models such as GPT-5~\cite{openai2025gpt5model} and Gemini~\cite{google2026gemini}, or train large open VLM backbones with long-horizon agentic reinforcement learning.

OpenSearch-VL~\cite{chen2026opensearchvl} changed the data landscape by releasing a 36K corpus of tool-interleaved teacher trajectories, an 8K RL prompt set, and a detailed seven-tool environment, but its strongest models are trained in a substantially larger regime than many research groups can access (agentic SFT on 256 H20 GPUs, then multi-turn GRPO on 64 H20 GPUs). This raises a separate question from frontier performance: \emph{what exactly do released multimodal search trajectories teach small VLMs?} Do they transfer factual search competence, or do they first teach the more basic agent contract---how to format tool calls, use image handles, integrate observations, and stop searching with a final answer?

We study this question in a deliberately constrained setting: starting from Qwen3-VL-2B and Qwen3-VL-4B, we train only LoRA adapters (vision tower frozen, multimodal projector trainable) on the released \dataset{} trajectories, construct a compact three-tool distilled variant to probe tool-surface complexity, and build synthetic step-level DPO pairs by asking GPT-5 to generate plausible local mistakes at teacher decision points.

Our evaluation covers ten model cells across four short-answer, knowledge-intensive benchmarks: SimpleVQA, FVQA, LiveVQA, and VDR-Bench-testmini. The clearest result is that SFT transfers the \emph{agent contract}: the 2B base model emits no usable final answer on 1{,}237 of 1{,}240 questions (0.1\% macro Pass@1), while after one epoch of full-trajectory SFT the same backbone reaches 28.4\% macro Pass@1 and matches the 4B base on the three easier benchmarks. This is not simply a scaling result: the 4B base already follows the contract on many questions, so its SFT gain is much smaller.

The later stages are more diagnostic than transformative: synthetic step-level DPO improves the macro average by +1.0 points on 2B and +1.7 points on 4B, with the largest localized gain on FVQA but a non-uniform effect across domains; the compact three-tool interface reduces performance at 2B, is nearly cost-free at 4B on average, and strongest on FVQA; and increasing the VDR step budget from 5 to 8 reduces \texttt{no\_answer} substantially but barely changes Pass@1---the recovered rollouts mostly become \texttt{wrong\_entity}, pointing at verification, not search depth, as the remaining bottleneck.

Our contributions are:
\begin{itemize}[leftmargin=*,itemsep=2pt]
  \item A low-compute recipe for 2B/4B multimodal search agents from released trajectories on one eight-GPU node.
  \item Full-trajectory SFT as a qualitative transition (2B \texttt{no\_answer} 1{,}237$\to$514; Pass@1 0.1\%$\to$28.4\%).
  \item Synthetic step-level DPO and compact tool distillation evaluated as refinements, not phase transitions.
  \item A failure-mode and step-budget analysis: VDR is limited by answer verification, not search turns.
\end{itemize}
We defer a full discussion of related work to Appendix~\ref{app:related}.

%--------------------------------------------------------------------------
\section{Setup and Method}
\label{sec:setup}

\methodname{} has three stages, described below after the agent formalism; Figure~\ref{fig:pipeline} (Appendix~\ref{app:floats}) gives an overview of the full pipeline.

\paragraph{Agent loop.}
\label{sec:prelim}
Given an image $I_0$ and a question $q$, a multimodal search agent produces interleaved reasoning, tool-use, and observation steps culminating in a final answer. At step $\ell$ the policy $\pi_\theta$ conditions on the history $h_\ell = (\mathcal{I}_\ell, q, \mathbf{a}_{<\ell}, \mathbf{o}_{<\ell})$, with $\mathcal{I}_\ell$ the growing visual context, and emits an action $a_\ell = [z_\ell, c_\ell]$: a chain-of-thought $z_\ell$ plus either a tool call $c_\ell \in \mathcal{T}$ or a final response, after which the deterministic environment $\mathcal{E}$ returns an observation $o_\ell = \mathcal{E}(c_\ell, h_\ell)$ that is appended to the history. A rollout terminates when the agent emits a \texttt{<response>} block or the step budget is reached. For a teacher trajectory $\tau = (h_0, a_0, o_0), \dots, (h_L, a_L)$ the standard supervised fine-tuning objective factors as
\begin{equation}
\label{eq:sft}
\mathcal{L}_{\text{SFT}}(\theta) = -\sum_{\ell=0}^{L} \left[\log P_\theta(z_\ell \mid h_\ell) + \log P_\theta(c_\ell \mid h_\ell, z_\ell)\right],
\end{equation}
with observation tokens masked from the loss, and for preference learning at a single step we follow DPO~\cite{rafailov2023dpo}:
\begin{equation}
\label{eq:dpo}
\mathcal{L}_{\text{DPO}}(\theta) = -\mathbb{E}_{(h, a^+, a^-)} \log \sigma\!\left(\beta \log \frac{\pi_\theta(a^+ \mid h)}{\pi_{\text{ref}}(a^+ \mid h)} - \beta \log \frac{\pi_\theta(a^- \mid h)}{\pi_{\text{ref}}(a^- \mid h)}\right),
\end{equation}
where $a^+$ is the teacher action and $a^-$ is a synthetic hard negative produced in Stage~C.

\paragraph{Tool environment.}
We faithfully re-implement the seven-tool environment of OpenSearch-VL, matching its prompt format, tool schema, and observation tokenization; Table~\ref{tab:tools} in Appendix~\ref{app:tools} lists every tool and backend. Retrieval tools are backed by SerpAPI~\cite{serpapi}; the visual-restoration and cropping tools are implemented in Pillow and OpenCV, and OCR uses docTR~\cite{doctr2021}. The dispatcher maintains an \texttt{img\_N} registry so that visual outputs of one tool can be referenced by index in subsequent calls, and every observation is wrapped in \texttt{<observation>...</observation>} matching the training-time tokenization exactly (required for stable inference).

\paragraph{Stage A: supervised distillation from released trajectories.}
\label{sec:method-sft}
We download the 7-domain \dataset{} corpus (per-domain statistics: Table~\ref{tab:data-stats}, Appendix~\ref{app:floats}) and apply three preprocessing passes: \emph{compression} strips the upstream chain-of-thought from tool observations and caps each cleaned observation at 4{,}000 characters, preserving the vast majority of observations end-to-end while removing roughly one third of total tokens; a deterministic per-trajectory hash produces a stratified 36{,}092 / 500 train / validation \emph{holdout split} with all seven domains represented in both sides; and \emph{step expansion} turns each $L$-turn trajectory into $L$ prefix$\to$next-step records, yielding 133{,}439 step-level examples (a 3.65$\times$ expansion). We fine-tune Qwen3-VL-2B and Qwen3-VL-4B on Eq.~\eqref{eq:sft} with LoRA adapters (rank $r=16$, $\alpha=32$) on the language tower, on a single eight-GPU node with mixed-precision DeepSpeed~\cite{rasley2020deepspeed} ZeRO-3; the 8{,}192-token cutoff covers the full trajectory distribution at our observation cap, and one epoch over the 36{,}092 train trajectories at effective batch size 64 corresponds to 564 optimizer steps (\texttt{2B SFT}, \texttt{4B SFT}). For a controlled comparison, \texttt{2B/4B Step-SFT} instead train on the prefix$\to$next-step expansion, capped to the same optimizer steps and effective batch size with all other hyperparameters identical, isolating the supervision unit from data quantity and optimization budget.

\paragraph{Stage B: tool-distilled SFT.}
\label{sec:method-distill}
Because the full seven-tool action space includes specialized image-restoration and OCR tools that are rarely invoked in the training distribution, we build a \emph{distilled} variant in which the assistant may only call \toolname{image\_search}, \toolname{text\_search}, and \toolname{crop}: we rewrite the system prompt and tool schema to be substantially more compact, simplify the \toolname{text\_search} argument set, and drop any trajectory (from the full 36{,}592-record corpus, train and validation combined) that used a disallowed tool, which removes 2{,}849 records (7.8\%) and leaves 33{,}743. Re-running the same deterministic per-domain stratified split on this filtered pool, targeting a fixed validation size of 500 for direct comparability with Stage A, yields 33{,}243 train and 500 validation trajectories; this fresh stratified validation sample overlaps the Stage A validation set on 450 records (50 of the original 500 are dropped by the tool whitelist and replaced by surviving records). Everything else (LoRA rank, optimizer, batch size) matches Stage A except for a shorter 6{,}144-token cutoff enabled by the compact prompt (Table~\ref{tab:hparams}), so the effective difference is the action space and prompt surface (\texttt{2B/4B 3-tool SFT}).

\paragraph{Stage C: synthetic step-level preference optimization.}
\label{sec:method-dpo}
Real on-policy rollouts at training time would dominate the budget; instead we exploit the $\sim$133K teacher decisions the preprocessed step-level corpus already provides. For each step we ask GPT-5 to synthesize \emph{hard negatives} that a competent but imperfect small VLM would plausibly emit at that decision point, conditioned on the image and the prior history, partitioned into five interpretable failure modes: \emph{premature answer} (a \texttt{<response>} before sufficient evidence), \emph{wrong tool} (structurally inappropriate for the next step), \emph{weak query} (right tool, unfocused or degenerate query), \emph{repeated query} (duplicates a search already in context), and \emph{ignored image} (ignores salient evidence in the registered \texttt{img\_N} images). Every candidate is filtered through a programmatic validator with five checks---format, mode-specific structural rules (e.g.\ a \emph{wrong\_tool} negative must change the tool), absence of answer leakage relative to the gold response, absence of meta-language in the reasoning trace, and consistency of image references with the running registry; after tightening the generation prompt with two additional rules (\emph{stay in character} and \emph{use only existing image handles}), the audited issue rate dropped from $54\%$ to $0\%$ on a fixed sample of records (Appendix~\ref{app:dpo-prompt}). Running the validated pipeline over the full 133K step corpus yields 324{,}121 accepted preference pairs from $\sim$96K teacher decisions (per-mode distribution in Figure~\ref{fig:dpo-modes}, Appendix~\ref{app:floats}); after dropping pairs beyond a length budget compatible with training memory (longer side $\le 5{,}120$ tokens), we retain $\sim$234K training and $\sim$3K validation pairs ($\sim$73\% of the full collection). Starting from each \texttt{SFT} checkpoint, we run DPO~\cite{rafailov2023dpo} with the sigmoid preference loss of Eq.~\eqref{eq:dpo} for one epoch over the filtered preference set at the SFT effective batch size (Table~\ref{tab:hparams}), using a relatively large $\beta=0.3$ together with a small learning rate of $10^{-6}$: the standard DPO recipe used on larger backbones (e.g.\ $\beta=0.1$ with a $5\!\times\!10^{-6}$ learning rate) collapsed the 2B model early in training, a related instability to the DPO failure mode documented in~\cite{pal2024smaug}. Final hyperparameters were selected by monitoring the reward gap on a held-out preference validation set (\texttt{2B/4B SFT+DPO}).

%--------------------------------------------------------------------------
\section{Evaluation Protocol}
\label{sec:eval}

\textbf{Benchmarks and inference.}\;
We evaluate on four short-answer, knowledge-intensive multimodal benchmarks: \textbf{FVQA}~\cite{wang2017fvqa} (59 questions from our held-out validation split, \texttt{fvqa} domain\footnote{The \texttt{fvqa} domain contributes 60 records to the validation split (Table~\ref{tab:data-stats}); one is excluded from the FVQA track because its image lacks a public URL required by the \toolname{image\_search} tool, leaving 59 evaluated questions. These are OpenSearch-VL trajectories built over FVQA~\cite{wang2017fvqa} images rather than items of the original FVQA test split, and are distinct from the FactualVQA set of MMSearch-R1~\cite{wu2025mmsearchr1}.}), \textbf{LiveVQA}~\cite{fu2025livevqa} (181 questions from our held-out validation split, \texttt{livevqa} domain), \textbf{SimpleVQA}~\cite{cheng2025simplevqa} (500 questions sampled from the EN subset of \texttt{m-a-p/SimpleVQA}~test), and \textbf{VDR}~\cite{zeng2026vdr} (500 questions from \texttt{Osilly/VDR-Bench-testmini}; VDR-Bench is the Vision-DeepResearch \emph{benchmark} of~\cite{zeng2026vdr}, distinct from the Vision-DeepResearch \emph{agent} of~\cite{huang2026visiondeepresearch}). Across the four benchmarks we evaluate ten model cells, each producing 1{,}240 agent rollouts for a total of 12{,}400 judged rollouts per evaluation pass. For each model we serve eight parallel single-GPU replicas with vLLM~\cite{kwon2023vllm} and dispatch rollouts round-robin with light concurrency; the per-rollout step budget is \texttt{MAX\_STEPS}~=~5 unless stated otherwise, and the input image is downscaled to a maximum edge of 360~px before being passed to the model. Tool calls to the search backend are deduplicated through a persistent on-disk cache so that the same (tool, arguments) pair is paid for at most once across all ten model cells in an evaluation pass, reducing the per-pass search-API cost by roughly an order of magnitude.

\textbf{LLM-as-judge and models compared.}\;
We use GPT-5 as an LLM-as-judge: it sees the original benchmark image, the question, the reference answer, and the model's final response, and must emit a strict JSON object with five fields: \texttt{correct}~$\in\{0,1\}$, \texttt{partially\_correct}~$\in\{0,1\}$ (mutually exclusive with \texttt{correct}=1), \texttt{error\_mode}~$\in$ \{\texttt{none}, \texttt{hallucinated}, \texttt{wrong\_entity}, \texttt{missing\_evidence}, \texttt{no\_answer}, \texttt{partial}, \texttt{off\_topic}, \texttt{other}\}, \texttt{confidence}~$\in\{0,1,2\}$, and a one-sentence \texttt{rationale} (full prompt in Appendix~\ref{app:judge-prompt}). The reported \emph{Pass@1} is the fraction with \texttt{correct}=1; the \emph{partial-credit rate} is $\text{Pass@1} + 0.5 \cdot \text{partial}$. All 12{,}400 main-evaluation judgments parse successfully as JSON, and the judge's self-rated \texttt{confidence} saturates at 2 on every judged row, so we treat it only as a pipeline-stability diagnostic (full discussion in Appendix~\ref{app:confidence}). Bootstrap confidence intervals over benchmark questions (Table~\ref{tab:pass1-ci}, Appendix~\ref{app:results-floats}) are our primary measure of result uncertainty; we recommend human spot-checking before treating any single judgment as ground truth. We compare ten model cells---the 2B and 4B variants of base, full-trajectory SFT, step-level SFT, 3-tool SFT, and SFT+DPO---but not the larger OpenSearch-VL 8B/30B/32B agents: the question we ask is whether 2B/4B agents can reach a useful operating point on a single eight-GPU node, not whether they overtake substantially larger frontier models (training-regime contrast: Table~\ref{tab:compute-footprint}, Appendix~\ref{app:floats}).

%--------------------------------------------------------------------------
\section{Results}
\label{sec:results}

Table~\ref{tab:main-pass1} reports Pass@1 across all ten model cells and four benchmarks, visualized in Figure~\ref{fig:main-results}; partial-credit rates (Table~\ref{tab:partial-credit}) and bootstrap confidence intervals for the macro averages (Table~\ref{tab:pass1-ci}) are deferred to Appendix~\ref{app:results-floats}.\footnote{Throughout the paper, stage-to-stage deltas (e.g.\ the +1.7-point DPO gain at 4B) are computed from unrounded scores before tabulation, so they can differ by 0.1 from differences of the rounded values printed in the tables.}

\begin{table}[t!]
% Main Pass@1 results table.
\centering
\caption{Pass@1 (\%) across the four short-answer benchmarks for ten model
cells. \textbf{Bold}: best in each column; \underline{underline}: best at 2B and 4B
separately; \textit{Avg}: macro-average over the four benchmarks. Cells are exact
$n_\text{correct} / n$ counts from the GPT-5 judge (Section~\ref{sec:eval}); $n=$
500, 59, 181, 500 for the four columns.}
\label{tab:main-pass1}
\small
\setlength{\tabcolsep}{6pt}
\begin{tabular}{l c c c c c}
\toprule
\textbf{Model} & \textbf{SimpleVQA} & \textbf{FVQA} & \textbf{LiveVQA} & \textbf{VDR} & \textbf{Avg} \\
\midrule
\multicolumn{6}{l}{\emph{Qwen3-VL-2B}} \\
\quad Base                              &  0.2 &  0.0 &  0.0 & 0.0 &  0.1 \\
\quad SFT (full traj.)                  & \underline{30.8} & 64.4 & 16.6 & \underline{2.0} & 28.4 \\
\quad Step-SFT                          & 24.2 & 61.0 & \underline{\textbf{21.0}} & 1.4 & 26.9 \\
\quad 3-tool SFT                        & 18.4 & 45.8 & 12.7 & 1.0 & 19.5 \\
\quad SFT + DPO                         & 30.6 & \underline{67.8} & 18.2 & 1.0 & \underline{29.4} \\
\midrule
\multicolumn{6}{l}{\emph{Qwen3-VL-4B}} \\
\quad Base                              & 24.6 & 61.0 & 16.0 & 0.6 & 25.6 \\
\quad SFT (full traj.)                  & 32.6 & 61.0 & \underline{19.9} & \underline{\textbf{2.6}} & 29.0 \\
\quad Step-SFT                          & \underline{\textbf{34.2}} & 66.1 & 18.8 & 1.2 & 30.1 \\
\quad 3-tool SFT                        & 21.0 & \underline{\textbf{71.2}} & 16.0 & 0.6 & 27.2 \\
\quad SFT + DPO                         & 34.0 & 69.5 & 17.1 & 2.4 & \underline{\textbf{30.8}} \\
\bottomrule
\end{tabular}

\vspace{8pt}

\centering
\includegraphics[width=0.5\linewidth]{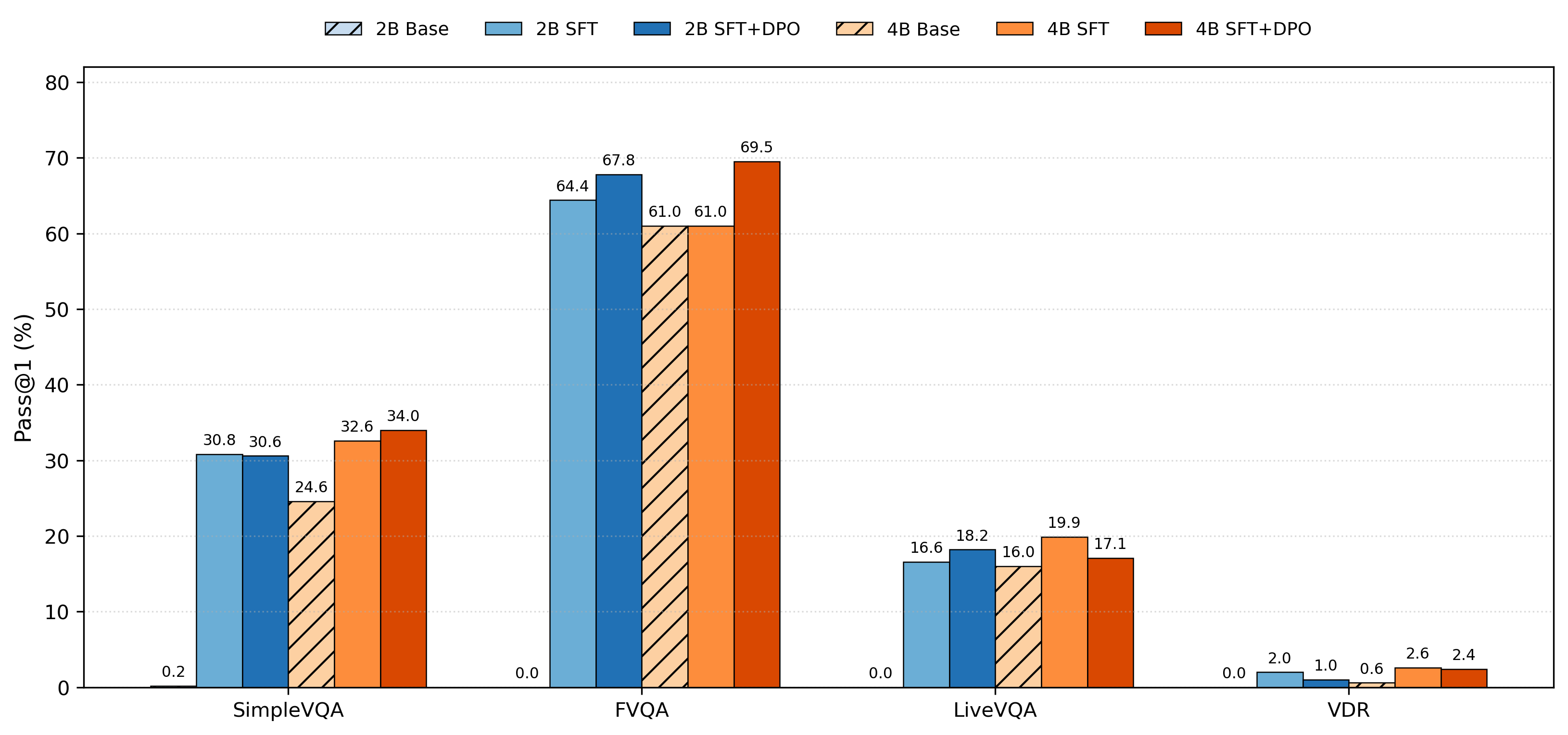}
\captionof{figure}{Pass@1 (\%) on the four benchmarks for the six headline model cells (the two step-SFT and the two 3-tool variants are omitted from the figure but listed in Table~\ref{tab:main-pass1}).}
\label{fig:main-results}
\end{table}

\textbf{(i) SFT transfers the agent contract, especially at 2B.}
The 2B base model fails primarily as an agent: it does not emit a usable \texttt{<response>} on 1{,}237 of 1{,}240 questions. Full-trajectory SFT cuts this failure mode to 514 cases and lifts macro Pass@1 to 28.4\%; the corresponding lift on 4B is much smaller (+3.5 points). After this transfer, the 2B SFT model reaches the level of the 4B base on FVQA (64.4\% vs.\ 61.0\%), SimpleVQA (30.8\% vs.\ 24.6\%), and LiveVQA (16.6\% vs.\ 16.0\%) (per-stage gains: Figure~\ref{fig:lift-summary}, Appendix~\ref{app:results-floats}).

\textbf{(ii) Synthetic step-level DPO is a refinement, not the phase transition.}
On top of full-trajectory SFT, the synthetic DPO pairs add +1.0 average points on 2B and +1.7 average points on 4B. The largest localized delta is on FVQA at 4B (61.0 $\to$ 69.5, +8.5 points, corresponding to five additional correct answers out of 59), while LiveVQA at 4B loses 2.8 points relative to SFT. The macro confidence intervals in Table~\ref{tab:pass1-ci} overlap, so we interpret DPO as a directional refinement that can reduce some local decision errors.

\textbf{(iii) Step-level SFT changes the terminal-behavior trade-off.}
On 2B, splitting the same trajectories into prefix$\to$next-step training units loses 1.5 average points relative to full-trajectory SFT and increases the \texttt{no\_answer} count from 514 to 780; on 4B the same change gains 1.0 points, despite increasing \texttt{no\_answer} from 587 to 682. Local transition supervision can help the larger backbone choose intermediate actions, but full-trajectory supervision more reliably teaches the small model the complete contract.

\textbf{(iv) A compact tool interface is not automatically easier.}
Restricting the action interface to three tools is worse on 2B (--9.0 average points) but nearly free on 4B (--1.8 average points) and strictly better on FVQA, where 4B 3-tool SFT is the strongest cell at 71.2\% Pass@1 (78.8\% partial-credit). The error-mode pattern is consistent with a capacity-dependent trade-off: the 3-tool 4B model commits fewer \texttt{wrong\_entity} mistakes than full-tool 4B SFT (242 vs.\ 298) but produces more \texttt{no\_answer} failures (766 vs.\ 587).

\textbf{Error modes.}
Table~\ref{tab:error-modes} and Figure~\ref{fig:error-modes} (Appendix~\ref{app:results-floats}) aggregate the GPT-5 judge's \texttt{error\_mode} labels across all 1{,}240 judged questions per cell. The dominant qualitative effect of SFT is collapsing the \texttt{no\_answer} pile of the base model into a roughly even split of \texttt{correct} and \texttt{wrong\_entity}; DPO further reduces \texttt{no\_answer} on 4B (587 $\to$ 511) and increases \texttt{correct} (248 $\to$ 254).

\textbf{Step-budget ablation on VDR.}
\label{sec:results-vdr}
VDR is the only benchmark in our suite where every model produces a large number of \texttt{no\_answer} rollouts at \texttt{MAX\_STEPS}=5 (59--70\% for the four SFT and DPO checkpoints re-run below; higher still for the base models). To test whether the step cap binds or is merely a symptom, we re-ran the four SFT and DPO models on VDR alone with \texttt{MAX\_STEPS}=8 (Table~\ref{tab:vdr-budget}; Figure~\ref{fig:vdr-budget}, Appendix~\ref{app:results-floats}). The extra step budget reduces \texttt{no\_answer} by roughly one third to one half on every model, but the additional rollouts mostly produce \texttt{wrong\_entity}, not \texttt{correct}: the bottleneck for small agents on VDR is therefore not simply search depth but evidence verification; Appendix~\ref{app:case-vdr-failure} traces one rollout step by step.

% VDR step-budget ablation table.
\begin{table}[t]
\centering
\caption{VDR step-budget ablation ($n=500$ each). Raising \texttt{MAX\_STEPS} from 5 to 8
cuts \texttt{no\_answer} by 33--50\% on every model; net Pass@1 moves within $\pm 1$ point,
and both DPO checkpoints improve (2B +0.8, 4B +1.0).}
\label{tab:vdr-budget}
\small
\setlength{\tabcolsep}{4pt}
\begin{tabular}{l c c c c c c c}
\toprule
\multirow{2}{*}{\textbf{Model}} & \multicolumn{2}{c}{\textbf{Pass@1 (\%)}} & \multicolumn{2}{c}{\textbf{Partial-credit (\%)}} & \multicolumn{2}{c}{\textbf{\texttt{no\_answer}}} & \multirow{2}{*}{\textbf{$\Delta$ Pass@1}} \\
\cmidrule(lr){2-3} \cmidrule(lr){4-5} \cmidrule(lr){6-7}
& 5 steps & 8 steps & 5 steps & 8 steps & 5 steps & 8 steps & \\
\midrule
2B SFT            & 2.0 & 2.4 & 5.2 & 7.5 & 304 & 171 & +0.4 \\
2B SFT + DPO      & 1.0 & 1.8 & 5.6 & 8.0 & 297 & 148 & +0.8 \\
4B SFT            & 2.6 & 2.4 & 6.4 & 8.3 & 351 & 236 & $-$0.2 \\
4B SFT + DPO      & 2.4 & 3.4 & 7.0 & 8.9 & 327 & 213 & +1.0 \\
\bottomrule
\end{tabular}
\end{table}

%--------------------------------------------------------------------------
\section{Discussion and Conclusion}
\label{sec:discussion}

\paragraph{Where SFT helps and where it doesn't.}
The 2B base model is essentially non-functional as a search agent, emitting a final \texttt{<response>} on only 3 of 1{,}240 questions (one of them correct); SFT repairs this, but the resulting agent still produces \texttt{wrong\_entity} on roughly 30\% of questions, so entity grounding and evidence checking remain unsolved. The 4B base already follows the contract on many questions; SFT mostly polishes its tool selection.

\paragraph{Assessment of synthetic DPO.}
The evidence supports a qualified version of the targeting claim: average Pass@1 grows on both backbones, with the cleanest single signal on 4B FVQA (+8.5 points) alongside a 13\% reduction in \texttt{no\_answer} over all four benchmarks (587 $\to$ 511, Table~\ref{tab:error-modes}). Conversely, LiveVQA at 4B loses 2.8 points after DPO, and the 2B picture is closer to a redistribution than a uniform improvement: 2B SFT and 2B+DPO have nearly identical partial-credit rates (35.4 in both rows of Table~\ref{tab:partial-credit}). We therefore frame DPO as a refinement on top of SFT, not a substitute, with real on-policy preferences the natural next step; Appendix~\ref{app:examples} walks through five paired case studies: the agent-contract transfer (\ref{app:case-base-vs-sft}), three DPO improvement mechanisms (\ref{app:case-2b-dpo}--\ref{app:case-4b-dpo-precision}), and a failure that survives DPO (\ref{app:case-vdr-failure}).

\paragraph{Tool space and backbone capacity.}
The 3-tool ablation crystallizes a capacity-dependent trade-off: the full tool set adds cognitive overhead, while the compact tool set removes specialized operations needed by some questions; larger backbones appear more robust to the former but still lose coverage when OCR or perspective correction is actually required---so tool-surface design should be treated as a model-capacity choice rather than a fixed property of the environment.

\paragraph{What the VDR ablation says about agent design.}
That three extra steps on VDR convert \texttt{no\_answer} into \texttt{wrong\_entity} rather than \texttt{correct} suggests that the limiting factor at small scales is not exploration but verification: the agent identifies the relevant region and retrieves plausible passages, but lacks a reliable habit of cross-checking the retrieved evidence against the question. A second-pass verification step, or a process reward model in the style of~\cite{lightman2024letsverify, setlur2024processreward}, is a natural direction.

\paragraph{Reproducibility.}
All ten model cells train with parameter-efficient LoRA on one eight-GPU node and are evaluated under the same agent environment and GPT-5 judge; we release training configurations, agent code, data pipelines, the judge prompt, and the search-cache schema (Appendix~\ref{app:repro}).

\paragraph{Conclusion.}
We presented \methodname{}, a low-compute recipe for adapting Qwen3-VL-2B/4B into multimodal search agents from released teacher trajectories. Across four short-answer benchmarks and 12{,}400 GPT-5-judged rollouts, the main transfer is the agent contract---full-trajectory SFT changes the 2B model from almost never answering to 28.4\% macro Pass@1---while compact tool distillation exposes a capacity-dependent trade-off and synthetic step-level DPO adds a smaller refinement (best 4B configuration: 30.8\% macro Pass@1 and 37.1\% partial-credit). The VDR ablation shows that extra search turns mainly convert abstentions into wrong-entity answers, pointing to verification-aware training as the next step.

%--------------------------------------------------------------------------
\clearpage
\bibliographystyle{unsrtnat}
\bibliography{references}

%--------------------------------------------------------------------------
\appendix

%==========================================================================
\section{Related Work}
\label{app:related}

\paragraph{Tool-augmented and agentic search.}
WebGPT~\cite{nakano2021webgpt} demonstrated browser-assisted question answering with human feedback, and ReAct~\cite{yao2023react} popularized the think-then-act prompting pattern that interleaves natural-language reasoning with discrete tool calls; Toolformer~\cite{schick2023toolformer} subsequently taught models to invoke tools in a self-supervised fashion. Search-R1~\cite{jin2025searchr1} introduced reinforcement learning with retrieval tokens masked out of the loss, and a number of multimodal search agents followed, including, in order of appearance, Visual-ARFT~\cite{liu2025visualarft}, MMSearch-R1~\cite{wu2025mmsearchr1}, WebWatcher~\cite{geng2025webwatcher}, DeepMMSearch-R1~\cite{narayan2025deepmmsearch}, DeepEyesV2~\cite{hong2025deepeyes}, SenseNova-MARS~\cite{chng2025sensenova}, and Vision-DeepResearch~\cite{huang2026visiondeepresearch}. The most directly related effort is OpenSearch-VL~\cite{chen2026opensearchvl}, which released a 36K supervised-finetuning corpus of multi-turn agent trajectories together with an 8K reinforcement-learning prompt set, and trained 8B--32B Qwen3-VL backbones with a fatal-aware variant of GRPO~\cite{shao2024deepseekmath}. Our work reuses the released data and tool schema, but targets a deliberately smaller and cheaper setting.

\paragraph{Parameter-efficient and preference-based post-training.}
LoRA~\cite{hu2022lora} and QLoRA~\cite{dettmers2023qlora} enable adapter-based fine-tuning of large backbones with a small fraction of trainable parameters. For preference alignment, DPO~\cite{rafailov2023dpo}, ORPO~\cite{hong2024orpo}, and SimPO~\cite{meng2024simpo} reduce the need for separate reward models, while Smaug~\cite{pal2024smaug} documents failure modes of DPO in which the likelihood of preferred completions declines during training. Step-DPO~\cite{lai2024stepdpo}, process reward models~\cite{lightman2024letsverify, setlur2024processreward}, and related work show that local, step-level preference signals can outperform trajectory-level rewards for long-horizon reasoning. Our synthetic step-level preference set adapts this perspective to multimodal agent rollouts, where each preference pair targets a single decision point inside a teacher trajectory.

\paragraph{Vision-language backbones and evaluation.}
We build on Qwen3-VL~\cite{bai2025qwen3vl} and the earlier Qwen2-VL~\cite{wang2024qwen2vl}; alternative backbones include LLaVA~\cite{liu2023llava}, InternVL~\cite{chen2024internvl}, and BLIP~\cite{li2022blip}. For evaluation we focus on benchmarks reported by OpenSearch-VL: FVQA~\cite{wang2017fvqa}, LiveVQA~\cite{fu2025livevqa}, SimpleVQA~\cite{cheng2025simplevqa}, VDR-Bench~\cite{zeng2026vdr}, MMSearch~\cite{jiang2024mmsearch}, and InfoSeek~\cite{chen2023infoseek}; WebQA~\cite{chang2022webqa} contributes to the training corpus rather than to that evaluation suite (Table~\ref{tab:data-stats}). We restrict ourselves to four short-answer benchmarks (FVQA, LiveVQA, SimpleVQA, VDR) that admit clean LLM-as-judge scoring and remain tractable at our compute budget. Training and serving use LlamaFactory~\cite{zheng2024llamafactory}, vLLM~\cite{kwon2023vllm}, and DeepSpeed~\cite{rasley2020deepspeed}.

%==========================================================================
\section{Pipeline Overview, Data Statistics, and Compute Footprint}
\label{app:floats}

This appendix collects the pipeline diagram, corpus statistics, synthetic-preference distribution, and compute-footprint comparison referenced from Sections~\ref{sec:setup} and~\ref{sec:eval}.

% TikZ pipeline diagram for LiteSearch-VL.
\begin{figure}[H]
\centering
\resizebox{\linewidth}{!}{%
\begin{tikzpicture}[
    font=\small,
    block/.style={rectangle, draw=black!60, rounded corners=2pt, align=center,
                  minimum height=10mm, minimum width=22mm, inner sep=4pt, fill=black!4},
    data/.style={cylinder, shape border rotate=90, draw=black!60, aspect=0.35,
                 minimum height=12mm, minimum width=24mm, inner sep=3pt,
                 fill=blue!8, align=center},
    model/.style={rectangle, draw=black!60, rounded corners=3pt, align=center,
                  minimum height=11mm, minimum width=26mm, inner sep=4pt,
                  fill=green!12, font=\small\bfseries},
    synth/.style={rectangle, draw=black!60, rounded corners=2pt, align=center,
                  minimum height=10mm, minimum width=22mm, inner sep=4pt,
                  fill=orange!12},
    eval/.style={rectangle, draw=black!60, rounded corners=2pt, align=center,
                 minimum height=10mm, minimum width=22mm, inner sep=4pt,
                 fill=violet!10},
    stage/.style={rectangle, draw=black!50, dashed, rounded corners=4pt, inner sep=8pt},
    arrow/.style={-Latex, draw=black!75, line width=0.6pt},
    initarrow/.style={-Latex, draw=black!60, line width=0.5pt, dashed},
    label/.style={font=\footnotesize\itshape, text=black!70, inner sep=1pt},
]

% ----------------------------------------------------------------
% Row 1: Data preparation (Stage A)
% ----------------------------------------------------------------
\node[data]  (sft36k)   at ( 0,    0) {SearchVL-SFT-36k\\\scriptsize 36{,}592 trajs};
\node[block] (compress) at ( 3.6,  0) {Compress\\\scriptsize +holdout};
\node[block] (split)    at ( 6.7,  0) {Split\\\scriptsize 36{,}092 / 500};
\node[block] (expand)   at ( 9.8,  0) {Step expand\\\scriptsize 133{,}439};

\draw[arrow] (sft36k)   -- (compress);
\draw[arrow] (compress) -- (split);
\draw[arrow] (split)    -- (expand);

\begin{scope}[on background layer]
  \node[stage, fit=(sft36k)(compress)(split)(expand),
        label={[label]above:Data preparation (Section~\ref{sec:method-sft})}] {};
\end{scope}

% ----------------------------------------------------------------
% Row 2: Training cells. SFT block is shifted left so the
% ``preferences'' arrow from Pref. pairs can travel up through clear
% space instead of crossing through the SFT box.
% ----------------------------------------------------------------
\node[model] (sftmodel)     at (-1.9, -3.6) {2B / 4B SFT};
\node[model] (distillmodel) at ( 1.0, -3.6) {2B / 4B\\3-tool SFT};
\node[model] (dpomodel)     at ( 4, -3.6) {2B / 4B SFT+DPO};

% Split -> SFT / 3-tool SFT (drops from Split.south down to a shared bus
% above the training row, then split into the two left destinations).
\draw[arrow] (split.south) -- ++(0, -0.6) -| (sftmodel.north);
\draw[arrow] (split.south) -- ++(0, -0.6) -| (distillmodel.north);

% SFT -> SFT+DPO initialization (arc above the row, clear of 3-tool box)
\draw[initarrow] (sftmodel.north)
    .. controls +(0,1.6) and +(0,1.6) ..
    node[label, above, midway, yshift=4pt] {initialize}
    (dpomodel.north);

% ----------------------------------------------------------------
% Row 3: Synthetic preference pipeline (right to left)
% ----------------------------------------------------------------
\node[synth] (gpt5)       at ( 6.2, -7.0) {GPT-5 vision\\\scriptsize negative synth};
\node[synth] (validators) at ( 2.0, -7.0) {5-layer\\validators};
\node[data]  (dpopairs)   at (-2.2, -7.0) {Pref. pairs\\\scriptsize 324{,}121};

% Step expand -> GPT-5 (long vertical drop along the right edge)
\draw[arrow] (expand.south) -- ++(0, -1.3) -|
   node[label, above, pos=0.25] {step records}
   (gpt5.north);

\draw[arrow] (gpt5)       -- (validators);
\draw[arrow] (validators) -- (dpopairs);

% Pref pairs -> SFT+DPO training: smooth curve through the empty band
% between row 2 and row 3, entering dpomodel from below. Avoids passing
% through any other model box.
\draw[arrow] (dpopairs.north)
    .. controls +(0,1.8) and +(-2.4,-1.6) ..
    node[label, above, pos=0.55] {preferences}
    (dpomodel.south);

\begin{scope}[on background layer]
  \node[stage,
        fit=(sftmodel)(distillmodel)(dpomodel)(dpopairs)(validators)(gpt5),
        inner sep=12pt,
        label={[label]below:Model training (Section~\ref{sec:setup}, Stages A--C)}] (trainstage) {};
\end{scope}

% ----------------------------------------------------------------
% Right column: evaluation
% ----------------------------------------------------------------
\node[eval] (env)     at (13.2, -2.4) {Agent rollouts\\\scriptsize 7-tool env + SerpAPI cache};
\node[eval] (judge)   at (13.2, -4.6) {GPT-5 judge\\\scriptsize 5-field rubric};
\node[data, fill=red!10] (results) at (13.2, -7.0)
   {Pass@1, partial,\\error modes};

% Trained checkpoints -> evaluation (from the whole training group: every
% checkpoint, not only SFT+DPO, is evaluated in the same environment)
\draw[arrow] (trainstage.east)
    -- node[label, above, sloped, pos=0.45] {all checkpoints}
    (env.west);

\draw[arrow] (env)   -- (judge);
\draw[arrow] (judge) -- (results);

\begin{scope}[on background layer]
  \node[stage, fit=(env)(judge)(results),
        inner sep=10pt,
        label={[label]above:Evaluation (Section~\ref{sec:eval})}] {};
\end{scope}

\end{tikzpicture}}
\caption{Overview of \methodname{}. The pipeline reuses the released
SearchVL-SFT-36k corpus to drive three model families
(full-trajectory SFT, 3-tool distilled SFT, and SFT+DPO with
mode-conditioned synthetic negatives). Every trained checkpoint is then
evaluated against the same seven-tool agent environment, with GPT-5
adjudicating each rollout under a fixed five-field rubric.}
\label{fig:pipeline}
\end{figure}
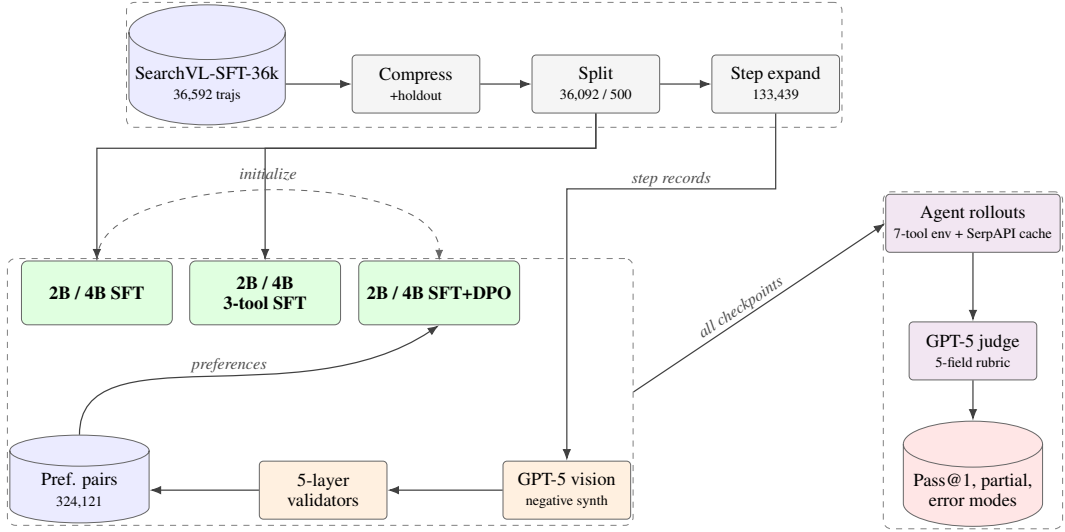

\begin{table}[H]
\centering
\caption{Composition of the \dataset{} corpus after preprocessing. \textit{Median turns} is the per-trajectory median across the train split.}
\label{tab:data-stats}
\small
\begin{tabular}{lrrrrr}
\toprule
Domain & Records & Train & Val & Step examples & Median turns \\
\midrule
livevqa  & 13{,}326 & 13{,}145 & 181 & 47{,}892 & 3 \\
wiki\_art &  5{,}093 &  5{,}023 &  70 & 19{,}027 & 4 \\
fvqa     &  4{,}413 &  4{,}353 &  60 & 14{,}628 & 3 \\
webqa    &  3{,}804 &  3{,}752 &  52 & 13{,}845 & 3 \\
wiki\_en  &  3{,}503 &  3{,}455 &  48 & 13{,}208 & 4 \\
wiki\_zh  &  3{,}480 &  3{,}432 &  48 & 13{,}210 & 4 \\
palace   &  2{,}973 &  2{,}932 &  41 & 11{,}629 & 4 \\
\midrule
\textbf{Total} & \textbf{36{,}592} & \textbf{36{,}092} & \textbf{500} & \textbf{133{,}439} & 3 \\
\bottomrule
\end{tabular}
\end{table}

\begin{figure}[H]
\centering
\includegraphics[width=0.72\linewidth]{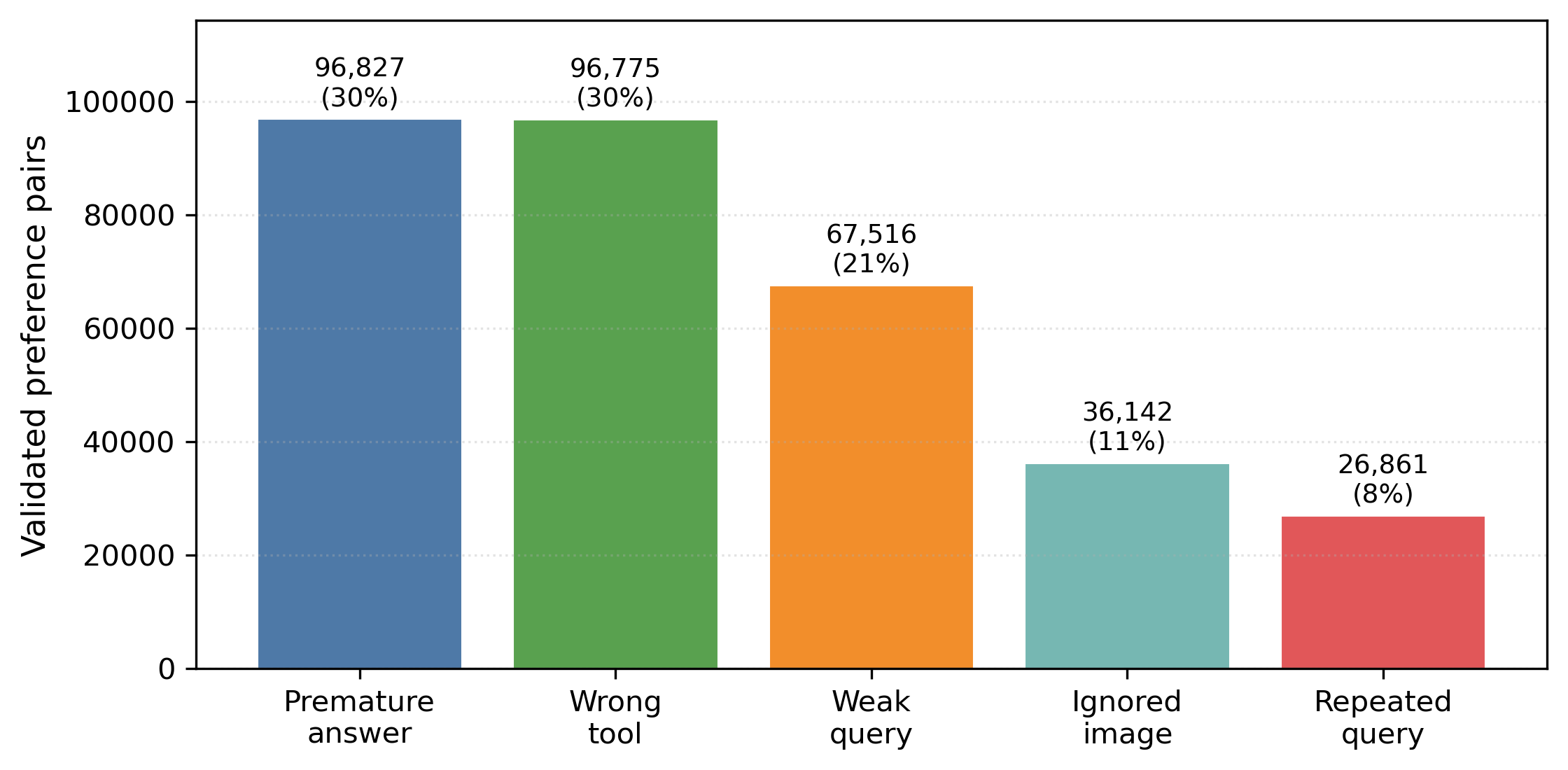}
\caption{Distribution of validated synthetic preference pairs across the five hard-negative failure modes. Each step record can contribute up to one pair per applicable mode.}
\label{fig:dpo-modes}
\end{figure}

\begin{table}[H]
\centering
\caption{Compute footprint relative to the OpenSearch-VL training recipe. OpenSearch-VL allocations are those reported in~\cite{chen2026opensearchvl}; \methodname{} reports the maximum concurrent training allocation used by our runs. The comparison is intended to clarify regime, not to normalize for hardware throughput.}
\label{tab:compute-footprint}
\small
\setlength{\tabcolsep}{4pt}
\begin{tabular}{p{0.17\linewidth} p{0.20\linewidth} p{0.27\linewidth} p{0.27\linewidth}}
\toprule
Recipe & Backbones & Training stages & Reported training allocation \\
\midrule
OpenSearch-VL & Qwen3-VL-8B / \mbox{30B-A3B} / 32B & Agentic SFT + multi-turn fatal-aware GRPO & SFT: 256 H20 GPUs for 2--4 days; GRPO: 64 H20 GPUs for 10 days \\
\methodname{} & Qwen3-VL-2B / 4B & LoRA SFT + synthetic step-level DPO; no on-policy RL & One eight-GPU node for each training stage \\
\bottomrule
\end{tabular}
\end{table}

%==========================================================================
\section{Judge Confidence Saturation}
\label{app:confidence}

We verified directly against the raw judge logs that \emph{every} judgment---all 12{,}400 main-evaluation rows, and all 14{,}400 rows once the four VDR step-budget ablation cells (Section~\ref{sec:results-vdr}) are included---reports \texttt{confidence}=2; the field never takes the value 0 or 1 anywhere in our data. We do not read this as evidence that every judgment is correct. Two properties of our setup make it plausible: (i) all four benchmarks use short factoid references (a name, date, or number), so deciding whether the model's response matches the reference is usually a clear-cut string/entity comparison rather than a graded call, which is precisely the case our rubric (Appendix~\ref{app:judge-prompt}) defines as \texttt{confidence}=2; and (ii) LLM self-rated confidence scores are known to saturate toward the top of a scale regardless of the rubric's intended granularity. Because \texttt{confidence} is not used to filter, reweight, or otherwise affect any reported Pass@1, partial-credit, or error-mode number, this saturation does not bias our headline results, but it does mean the field carries no discriminative signal in our evaluation and should be read as a pipeline-stability diagnostic (JSON parses, a value is always emitted) rather than an independent reliability check.

%==========================================================================
\section{Additional Results Tables and Figures}
\label{app:results-floats}

\begin{figure}[H]
\centering
\begin{subfigure}[t]{0.49\linewidth}
\centering
\includegraphics[width=\linewidth]{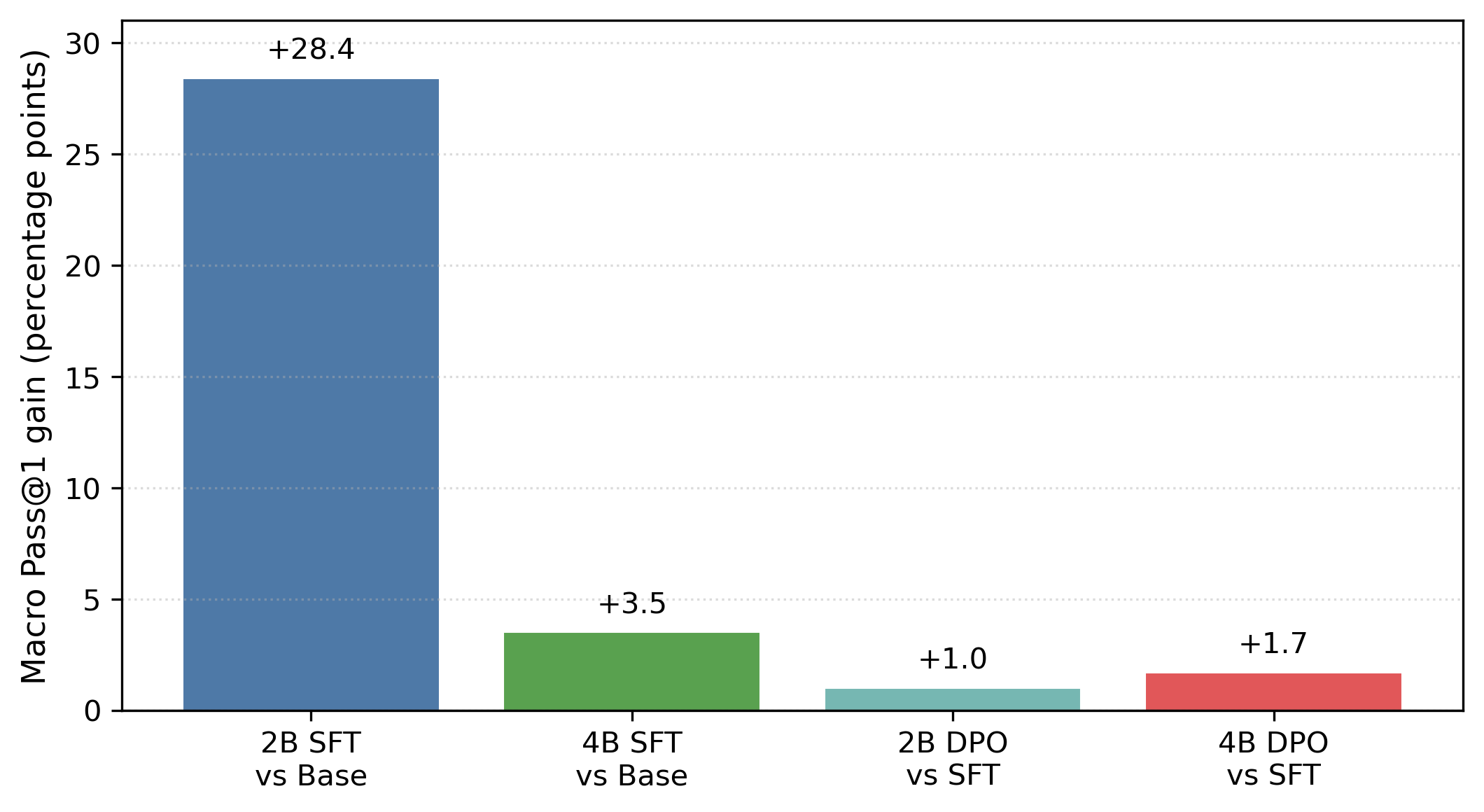}
\caption{Average Pass@1 gain in percentage points across the four benchmarks. SFT contributes most of the improvement, especially on 2B.}
\label{fig:lift-summary}
\end{subfigure}\hfill
\begin{subfigure}[t]{0.49\linewidth}
\centering
\includegraphics[width=\linewidth]{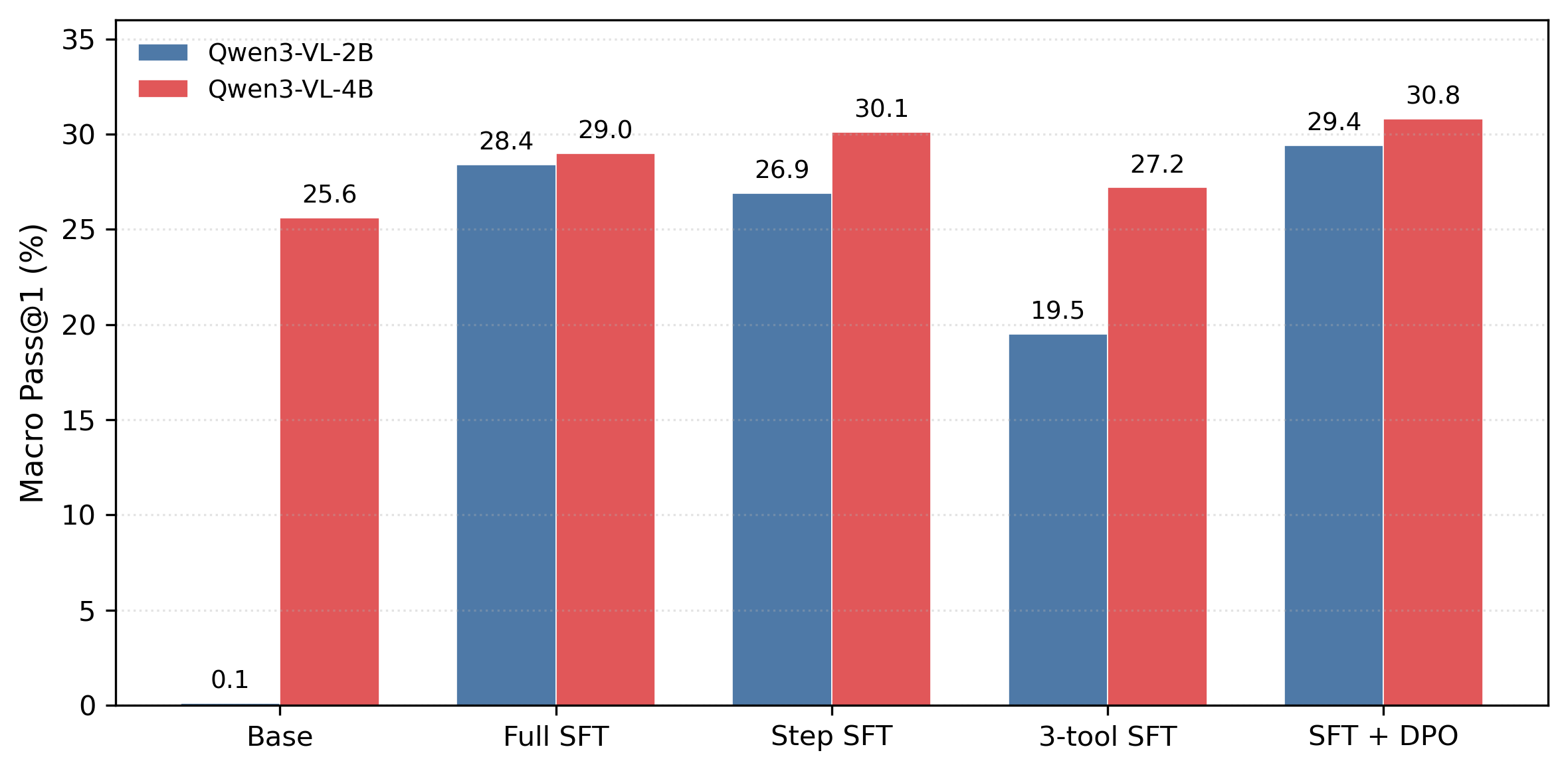}
\caption{Macro Pass@1 by backbone size and training regime. Within the small-model regime, the 4B+DPO checkpoint is the strongest configuration.}
\label{fig:size-vs-regime}
\end{subfigure}
\caption{Two views of the main result: the absolute gain attributable to each training stage (left), and the average Pass@1 across regimes for the 2B and 4B backbones (right).}
\label{fig:lift-and-size}
\end{figure}

\begin{figure}[H]
\centering
\includegraphics[width=0.92\linewidth]{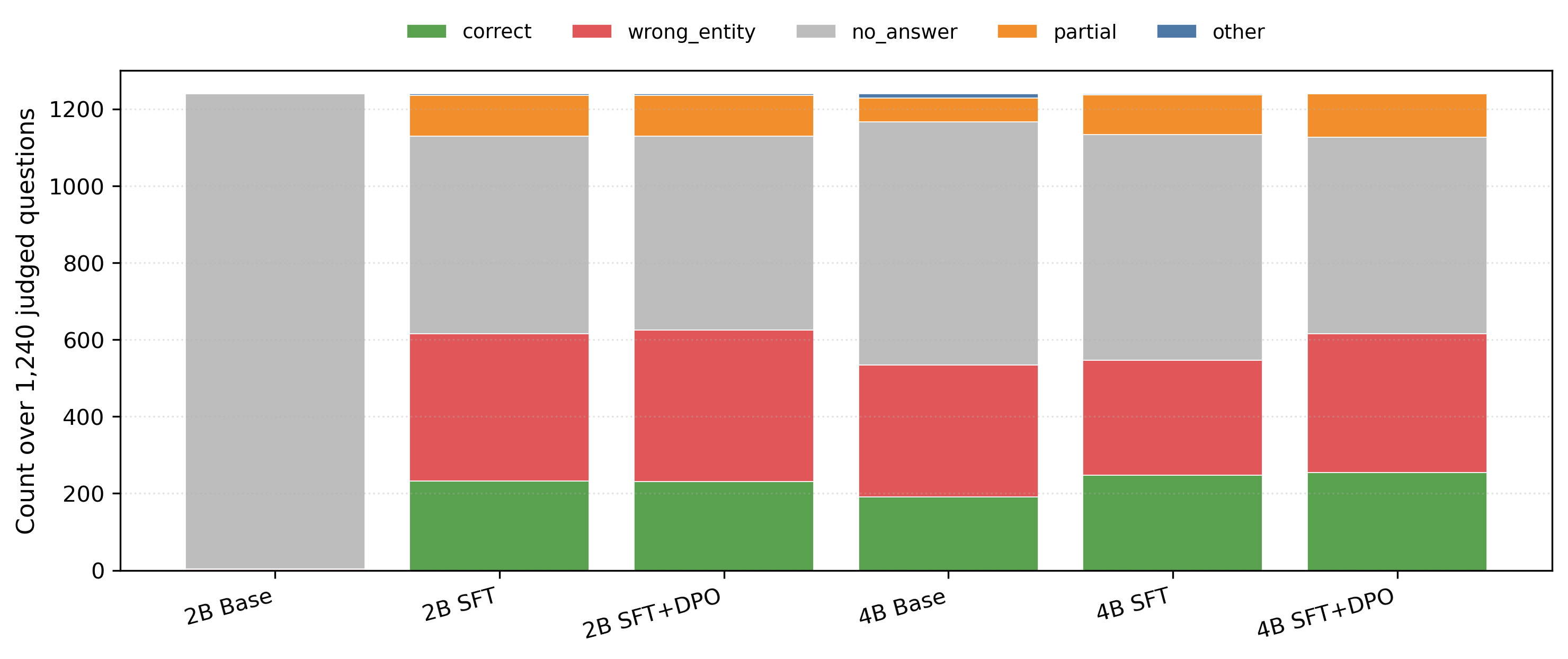}
\caption{Stacked error-mode distribution per model. The dominant qualitative effect of SFT is collapsing the \texttt{no\_answer} pile of the base model into a roughly even split of \texttt{correct} and \texttt{wrong\_entity}. DPO further reduces \texttt{no\_answer} on 4B (587 $\to$ 511) and increases \texttt{correct} (248 $\to$ 254).}
\label{fig:error-modes}
\end{figure}

\begin{figure}[H]
\centering
\includegraphics[width=0.78\linewidth]{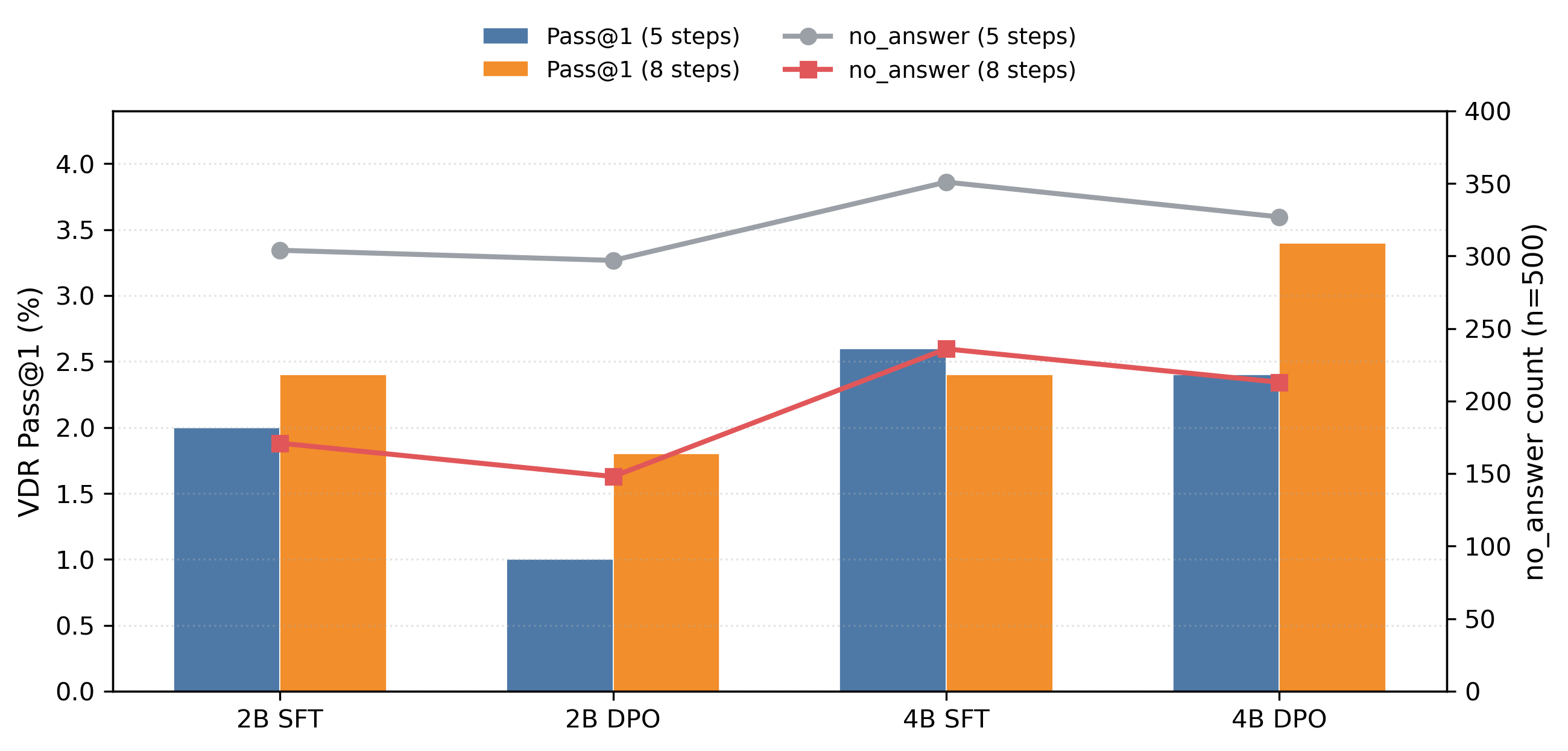}
\caption{VDR step-budget ablation. The extra step budget reduces \texttt{no\_answer} by roughly one third to one half on every model (right axis, solid lines), but the additional rollouts mostly produce \texttt{wrong\_entity}, not \texttt{correct}. Net Pass@1 movement is within $\pm 1$ point on all four models.}
\label{fig:vdr-budget}
\end{figure}

Table~\ref{tab:partial-credit} reports partial-credit rates for all ten model cells, Table~\ref{tab:error-modes} the full error-mode distributions, Table~\ref{tab:pass1-ci} the bootstrap confidence intervals for the macro averages, Figure~\ref{fig:lift-and-size} the per-stage gains (Figure~\ref{fig:lift-summary}) and the macro averages across all five regimes for both backbones (Figure~\ref{fig:size-vs-regime}), Figure~\ref{fig:error-modes} the stacked error-mode distributions, and Figure~\ref{fig:vdr-budget} the VDR step-budget ablation, all deferred from Section~\ref{sec:results}.

% Partial-credit results table.
\begin{table}[H]
\centering
\caption{Partial-credit rate (\%), computed as $\text{Pass@1} + 0.5\cdot\text{partial}$, captures
``right entity, wrong sub-attribute'' answers that strict Pass@1 misses. The
relative ranking of models is similar, but the 4B+DPO checkpoint is the clear
winner on the macro average. \emph{Note:} the implied partial-response counts
from this table (via $2\times(\text{partial-credit}-\text{Pass@1})\times n$)
can differ from the discrete \texttt{error\_mode=partial} tally in
Table~\ref{tab:error-modes} by exactly one row for two cells (2B 3-tool SFT:
68 vs.\ 67; 4B SFT: 105 vs.\ 104). This traces to judged rows where GPT-5
set \texttt{partially\_correct=1} but labeled \texttt{error\_mode} as
something else (e.g.\ \texttt{hallucinated}, \texttt{wrong\_entity}),
breaking the intended mutual exclusivity between the two signals: two of the
12{,}400 main-evaluation rows (one per affected cell), plus a third such row
in the VDR step-budget ablation of Section~\ref{sec:results-vdr} (3 of all
14{,}400 judgments). It does not affect Pass@1, partial-credit rates, or any reported
ranking.}
\label{tab:partial-credit}
\small
\setlength{\tabcolsep}{6pt}
\begin{tabular}{l c c c c c}
\toprule
\textbf{Model} & \textbf{SimpleVQA} & \textbf{FVQA} & \textbf{LiveVQA} & \textbf{VDR} & \textbf{Avg} \\
\midrule
2B Base                  &  0.2 &  0.0 &  0.0 & 0.0 &  0.1 \\
2B SFT                   & 33.7 & 78.0 & 24.6 & 5.2 & 35.4 \\
2B Step-SFT              & 26.4 & 72.9 & 24.6 & 3.8 & 31.9 \\
2B 3-tool SFT            & 20.5 & 54.2 & 16.6 & 3.3 & 23.7 \\
2B SFT + DPO             & 32.7 & 78.0 & 25.4 & 5.6 & 35.4 \\
\midrule
4B Base                  & 26.2 & 64.4 & 20.7 & 3.1 & 28.6 \\
4B SFT                   & 34.9 & 76.3 & 27.1 & 6.4 & 36.2 \\
4B Step-SFT              & 36.2 & 77.1 & 23.2 & 4.3 & 35.2 \\
4B 3-tool SFT            & 22.6 & 78.8 & 18.8 & 2.4 & 30.6 \\
4B SFT + DPO             & 36.5 & 79.7 & 25.1 & 7.0 & 37.1 \\
\bottomrule
\end{tabular}
\end{table}

% Error-mode distribution table.
\begin{table}[H]
\centering
\caption{Error-mode distribution per model. Counts are over all
$n=1{,}240$ judged questions (sum of the four benchmarks). \texttt{none}
denotes the correct bucket, \texttt{partial} is the partial-credit bucket,
and the other modes are mutually exclusive failure types from the GPT-5
rubric (Section~\ref{sec:eval}). The 2B base model produces almost no
final answers, which is the failure pattern SFT primarily addresses.
\emph{Note:} the \texttt{partial} column counts rows with the discrete label
\texttt{error\_mode=partial}, whereas Table~\ref{tab:partial-credit} is
computed from the \texttt{partially\_correct} flag; for 2B 3-tool
SFT and 4B SFT one judged row each sets \texttt{partially\_correct=1} while
labeling \texttt{error\_mode} differently (two of the 12{,}400
main-evaluation rows; a third such row occurs in the VDR step-budget
ablation), so those two rows sum to one less than
Table~\ref{tab:partial-credit} implies
(67 vs.\ 68 and 104 vs.\ 105, respectively). See the note on
Table~\ref{tab:partial-credit}.}
\label{tab:error-modes}
\small
\setlength{\tabcolsep}{6pt}
\begin{tabular}{l r r r r r}
\toprule
\textbf{Model} & \textbf{\texttt{none} $\uparrow$} & \textbf{\texttt{wrong\_entity}} & \textbf{\texttt{no\_answer}} & \textbf{\texttt{partial}} & \textbf{other} \\
\midrule
2B Base                  &     1 &   2 & 1{,}237 &    0 &  0 \\
2B SFT                   &   232 & 383 &   514 &  106 &  5 \\
2B Step-SFT              &   202 & 184 &   780 &   73 &  1 \\
2B 3-tool SFT            &   147 & 450 &   572 &   67 &  4 \\
2B SFT + DPO             &   231 & 394 &   505 &  105 &  5 \\
\midrule
4B Base                  &   191 & 343 &   632 &   62 & 12 \\
4B SFT                   &   248 & 298 &   587 &  104 &  3 \\
4B Step-SFT              &   250 & 225 &   682 &   80 &  3 \\
4B 3-tool SFT            &   179 & 242 &   766 &   53 &  0 \\
4B SFT + DPO             & 254 & 362 &   511 &  112 &  1 \\
\bottomrule
\end{tabular}
\end{table}

\begin{table}[H]
\centering
\caption{Macro Pass@1 uncertainty from the existing question-level GPT-5 judgments. For each model, we bootstrap questions within each benchmark independently and recompute the four-benchmark macro average; intervals are percentile 95\% CIs over 10{,}000 bootstrap replicates.}
\label{tab:pass1-ci}
\small
\setlength{\tabcolsep}{6pt}
\begin{tabular}{lcc}
\toprule
Model & Avg Pass@1 (\%) & 95\% CI (\%) \\
\midrule
2B Base & 0.1 & [0.0, 0.1] \\
2B SFT & 28.4 & [24.9, 31.9] \\
2B Step-SFT & 26.9 & [23.3, 30.5] \\
2B 3-tool SFT & 19.5 & [16.0, 23.0] \\
2B SFT + DPO & 29.4 & [26.0, 32.8] \\
\midrule
4B Base & 25.6 & [22.1, 29.0] \\
4B SFT & 29.0 & [25.4, 32.5] \\
4B Step-SFT & 30.1 & [26.6, 33.5] \\
4B 3-tool SFT & 27.2 & [23.8, 30.4] \\
4B SFT + DPO & 30.8 & [27.4, 34.1] \\
\bottomrule
\end{tabular}
\end{table}

%==========================================================================
% Existing arXiv appendix (hyperparameters, tool environment, judge prompt,
% preference generator, per-domain statistics, reproducibility checklist,
% qualitative case studies) -- content unchanged.
% Appendix for LiteSearch-VL.

\section{Hyperparameters and Compute}
\label{app:hparams}

Table~\ref{tab:hparams} lists the training hyperparameters for every
training run; its five rows cover four distinct configurations, since the
2B and 4B SFT runs differ only in backbone. SFT runs on a single eight-GPU node with mixed-precision
DeepSpeed ZeRO-3, and DPO runs on a single eight-GPU node with ZeRO-2.
Backbones, LoRA configuration, and effective batch size are kept
identical across stages to enable clean comparisons.

\begin{table}[h]
\centering
\caption{Training hyperparameters. ``$L_{\max}$'' is the model context
length in tokens, ``$B_{\text{eff}}$'' is the effective batch size, and
``Steps'' is the total number of optimizer steps in one epoch.}
\label{tab:hparams}
\small
\setlength{\tabcolsep}{5pt}
\begin{tabular}{l l l l c c c c}
\toprule
\textbf{Stage} & \textbf{Data} & \textbf{Adapter} & \textbf{Precision} & $L_{\max}$ & LR & $B_{\text{eff}}$ & Steps \\
\midrule
2B SFT             & $\sim$36K traj.   & LoRA-16 & fp16 & 8{,}192 & $1\!\times\!10^{-4}$ & 64 &  564 \\
4B SFT             & $\sim$36K traj.   & LoRA-16 & fp16 & 8{,}192 & $1\!\times\!10^{-4}$ & 64 &  564 \\
2B/4B Step-SFT     & $\sim$36K steps   & LoRA-16 & fp16 & 8{,}192 & $1\!\times\!10^{-4}$ & 64 &  564 \\
2B/4B 3-tool SFT   & $\sim$33K traj.   & LoRA-16 & fp16 & 6{,}144 & $1\!\times\!10^{-4}$ & 64 &  519 \\
2B/4B DPO          & $\sim$234K pairs  & LoRA-16 & bf16 & 5{,}120 & $1\!\times\!10^{-6}$ & 64 & 3{,}663 \\
\bottomrule
\end{tabular}
\end{table}

All SFT runs use LoRA with rank $r=16$ and scale $\alpha=32$, dropout
$0.05$, applied to all linear projections of the language tower. The
vision tower is frozen; the multimodal projector is trainable. We use
cosine learning-rate schedules with 3\% warmup, gradient checkpointing,
and image inputs are clipped to a maximum of 65{,}536 pixels
(approximately $256\times 256$) during training.

For DPO we use the sigmoid preference loss with $\beta=0.3$ and a small
learning rate of $10^{-6}$; the standard recipe ($\beta=0.1$, learning rate
$5\!\times\!10^{-6}$) was unstable on the 2B backbone, a related instability to the DPO
failure mode discussed in~\cite{pal2024smaug}. The
reference model is the corresponding SFT checkpoint.

\section{Tool Environment}
\label{app:tools}

Table~\ref{tab:tools} summarizes the seven distinct tools exposed to the
agent; \toolname{web\_search} is dispatched as an alias of
\toolname{text\_search}, giving eight callable names.
The full tool schema and JSON argument signatures are released with the
code. The dispatcher maintains an \texttt{img\_N} registry across turns;
visual tools return a new \texttt{img\_N+1} reference and the original
training format wraps every observation in
\texttt{<observation>Tool execution result:\textbackslash n...\textbackslash n</observation>}.

\begin{table}[h]
\centering
\caption{The agent's tool environment at inference time (seven distinct
tools; \toolname{web\_search} is an alias of \toolname{text\_search}). The 3-tool
distilled variant (Section~\ref{sec:method-distill}) restricts the
action space to \toolname{image\_search}, \toolname{text\_search}, and
\toolname{crop}.}
\label{tab:tools}
\small
\begin{tabular}{l l l}
\toprule
\textbf{Tool} & \textbf{Backend} & \textbf{Observation format} \\
\midrule
\toolname{text\_search}        & SerpAPI \texttt{google}      & Passage blocks (title / URL / summary) \\
\toolname{web\_search}         & alias of \toolname{text\_search} & same as \toolname{text\_search} \\
\toolname{image\_search}       & SerpAPI \texttt{google\_lens} & Python-repr list of \texttt{\{title, source\}} \\
\toolname{crop}                & Pillow                       & New image id + storage URL \\
\toolname{layout\_parsing}     & docTR OCR                    & Text blocks + recognized text \\
\toolname{sharpen}             & Pillow unsharp mask          & Same template as \toolname{crop} \\
\toolname{super\_resolution}   & Bicubic 2$\times$ upscaling  & Same template as \toolname{crop} \\
\toolname{perspective\_correct} & OpenCV contour + warp       & Same template as \toolname{crop} \\
\bottomrule
\end{tabular}
\end{table}

\section{GPT-5 Judge Prompt}
\label{app:judge-prompt}

The judge receives the benchmark image, the question, the reference
answer, and the model response. The system rubric is reproduced
verbatim below.

\begin{lstlisting}[style=prompt]
You are a strict but fair grader for a multimodal question-answering
benchmark. You will see the original image (above this prompt), a
QUESTION, a REFERENCE answer (ground truth), and a MODEL response.
Compare the MODEL response to the REFERENCE answer for that QUESTION
about that IMAGE, and emit a structured JSON judgment with five fields.

Scoring rules:
1. `correct = 1` iff the MODEL response gives the SAME substantive
   answer as the REFERENCE. Tolerate phrasing, length, capitalization,
   formatting differences. Extra background prose is fine if the core
   answer is present and consistent.
2. `correct = 0` if the model's answer is missing, contradicts the
   reference, identifies the wrong entity/date/number, or is so vague
   it cannot be matched.
3. `partially_correct = 1` iff `correct = 0` AND the model got the
   primary entity / key subject right but missed or contradicted a
   secondary attribute (e.g. correct person but wrong year; correct
   landmark but wrong height). Otherwise 0. If `correct = 1`, set 0.
4. For factoid answers (names, dates, numbers, places): the substantive
   token must appear and must agree with the REFERENCE (allow obvious
   spelling / transliteration variants).
5. `error_mode` is one of: 'none' (when correct=1), 'hallucinated'
   (fabricated facts), 'wrong_entity' (named wrong subject),
   'missing_evidence' (plausible answer but no real grounding),
   'no_answer' (empty / refused / truncated before answer), 'partial'
   (got main entity but missed a sub-detail - pair with
   partially_correct=1), 'off_topic' (answered a different question),
   'other'.
6. `confidence` is your own confidence in this judgment: 2 = obvious,
   1 = required some interpretation, 0 = genuinely ambiguous.
7. Ignore the model's reasoning trace, tool calls, and markup - only
   the final answer content matters.
8. Output STRICT JSON ONLY, no markdown fences, no extra text, exactly:
   {"correct": 0|1, "partially_correct": 0|1,
    "error_mode": "<one of the 8 modes>", "confidence": 0|1|2,
    "rationale": "<one sentence>"}
\end{lstlisting}

The per-item prompt then appends the question, reference, and truncated
model response in three labeled sections and ends with the directive
\texttt{Return STRICT JSON only}.

\section{Synthetic Preference Generator}
\label{app:dpo-prompt}

For each step record in the corpus, the generator receives the input
image, the conversation prefix, the teacher's actual next step (the
\emph{chosen} action), and the gold final answer (used only for leak
detection). For each applicable failure mode it asks GPT-5 to produce a
plausible, hard, \emph{in-character} alternative action that a competent
but imperfect small VLM might emit. The system prompt enforces eight
hard rules:

\begin{lstlisting}[style=prompt]
1. Stay in character. The thought must read as if the model believes
   the chosen action is correct. Do NOT betray the mistake with
   meta-language such as "even though", "instead of doing the right
   thing", "I might mistakenly", "for this exercise".
2. Use only existing image handles. Image-consuming tools must
   reference a `img_N` that appears in the prior history. Do NOT
   fabricate filenames or URLs.
3. Respect the tool schema. The tool name must be one of the allowed
   tools; arguments must match the JSON schema.
4. Match the requested failure mode structurally. For example, a
   `wrong_tool` action must change the tool name vs. the chosen
   action; a `weak_query` action must keep the same tool; a
   `repeated_query` action must share at least one bag-of-words
   match with a prior call.
5. Do NOT leak the gold answer. The rejected action must not contain
   the gold final answer string or any obvious paraphrase.
6. Keep the format. The action must be a contiguous
   <think>...</think><tool_call>{...}</tool_call> block, or
   <think>...</think><response>...</response> for the
   `premature_answer` mode.
7. Be hard. The rejected action should be the kind of mistake a real
   small model would make, not a strawman.
8. Output strict JSON keyed by mode; one entry per applicable mode.
\end{lstlisting}

A programmatic validator (format / mode-specific rules / answer leak /
meta-language leak / image argument sanity) drops any entry that violates
a rule before it is written to the final pair file. After adding the
\emph{stay in character} and \emph{existing image handles only} rules,
the audited issue rate dropped from $54\%$ to $0\%$ on a fixed sample of
records (Section~\ref{sec:method-dpo}).

\section{Per-Domain Validation Statistics}
\label{app:per-domain}

For completeness, Table~\ref{tab:val-domain} reports the per-domain
composition of the held-out validation set, including the two domains
(\texttt{fvqa}, \texttt{livevqa}) that we promote to standalone
benchmark tracks.

\begin{table}[h]
\centering
\caption{Per-domain composition of the held-out 500-record validation
split. The \texttt{fvqa} and \texttt{livevqa} subsets are used as the
two corresponding benchmark tracks in the main evaluation; the other
five domains are kept for ablation but not reported in the main table.
The FVQA track evaluates 59 of the 60 \texttt{fvqa} records: one record
is excluded because its image lacks a public URL required by
\texttt{image\_search}.}
\label{tab:val-domain}
\small
\begin{tabular}{l r r}
\toprule
\textbf{Domain} & \textbf{Val records} & \textbf{Used as benchmark?} \\
\midrule
livevqa  & 181 & yes (LiveVQA track) \\
wiki\_art &  70 & no \\
fvqa     &  60 & yes (FVQA track) \\
webqa    &  52 & no \\
wiki\_en  &  48 & no \\
wiki\_zh  &  48 & no \\
palace   &  41 & no \\
\midrule
\textbf{Total} & \textbf{500} & \\
\bottomrule
\end{tabular}
\end{table}

\section{Reproducibility Checklist}
\label{app:repro}

\begin{itemize}[leftmargin=*,itemsep=2pt]
\item \textbf{Code and configs.} All ten model cells in
    Table~\ref{tab:main-pass1} are produced by configuration files that
    share a common template; the only differences are the base model,
    the dataset, and the training stage.
\item \textbf{Data pipeline.} The preprocessing script rebuilds the
    train / validation split and the step-level expansion deterministically
    given a fixed seed.
\item \textbf{Tool environment.} The agent loop, tool dispatcher, and
    search cache are shared across the interactive playground and the
    evaluation harness, ensuring identical inference behavior.
\item \textbf{Judge.} The GPT-5 judge prompt is fixed
    (Appendix~\ref{app:judge-prompt}) and applied identically to every
    model; the judge's self-rated \texttt{confidence} field is 2 (the
    maximum) on every individual judged row, not merely on average per
    cell. We do not treat this as a reliability guarantee --- see the
    discussion in Section~\ref{sec:eval} for why we believe this reflects
    the clear-cut nature of factoid matching in our benchmarks rather than
    a well-calibrated confidence signal.
\item \textbf{Caching.} Search-API results are cached on disk and shared
    across all ten model cells in a single evaluation pass; error-shaped
    observations are not cached.
\end{itemize}

\section{Qualitative Case Studies}
\label{app:examples}

This appendix presents five rollouts drawn directly from the evaluation
harness to illustrate, on real questions, what each training stage
changes about agent behavior. Each case includes the benchmark
question, the gold reference answer, a brief image description, the
trajectory of each model side by side, the model's final answer, and
the GPT-5 judge's verdict. Trajectories are excerpted from
\texttt{eval\_outputs/<model>/<benchmark>/responses.jsonl}; long
\texttt{<think>} blocks, tool observations, and intermediate tool calls
are elided with ``\textit{[\ldots]}'' for readability, and the text
shown is otherwise unmodified. The
first four cases study \emph{stage-wise progress} (Base $\to$ SFT, and
three mechanisms by which DPO improves over SFT at 2B and 4B); the final
case studies a \emph{failure mode that survives DPO}: the verification
bottleneck on VDR (Section~\ref{sec:results-vdr}).

%--------------------------------------------------------------------------
\subsection{Case 1: SFT teaches the agent contract (2B Base $\to$ 2B SFT)}
\label{app:case-base-vs-sft}

\begin{figure}[h]
\centering
\includegraphics[width=0.32\linewidth]{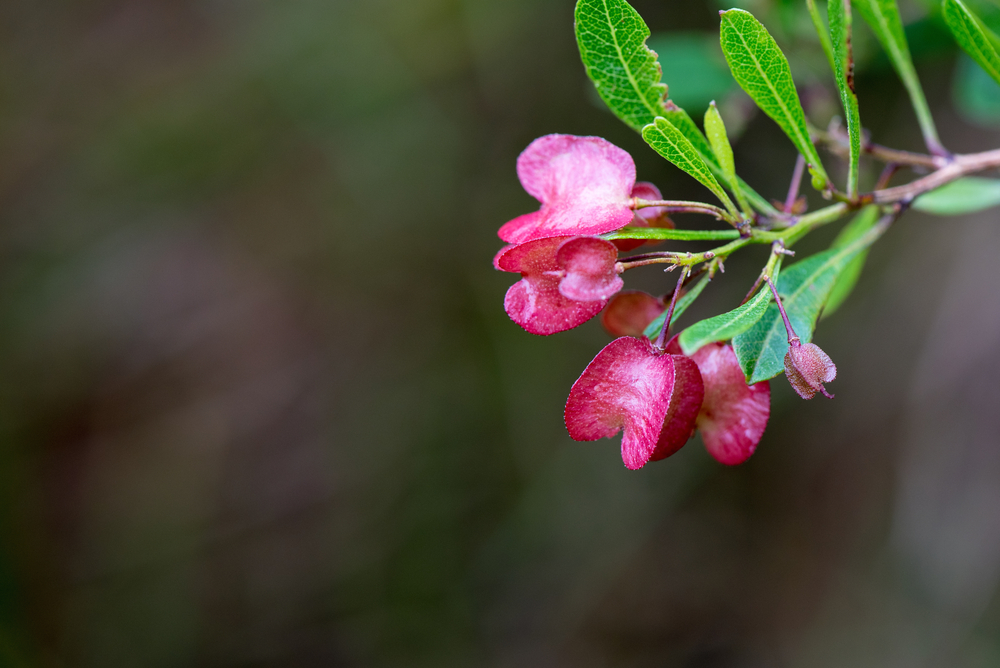}
\caption{Image for Case~\ref{app:case-base-vs-sft} (FVQA
\texttt{fvqa\_002606}).}
\label{fig:case1-img}
\end{figure}

\paragraph{Setup.} Benchmark: FVQA, item \texttt{fvqa\_002606}. The
image shows a branch of an ornamental shrub bearing clusters of
small, pink, winged seed pods against a blurred green background (Figure~\ref{fig:case1-img}).

\paragraph{Question.} ``What is the name of this plant with pink
seed pods?''

\paragraph{Reference.} \emph{Dodonaea viscosa (commonly known as
hop-bush or hopseed bush).}

\paragraph{Qwen3-VL-2B Base.} The base model executes the right tools
in the right order and even drafts a correct identification sentence,
but never wraps its answer in a \texttt{<response>} block, so the
harness records \texttt{no\_answer}.
\begin{lstlisting}[style=prompt]
[user]        <image>What is the name of this plant with pink seed pods?
[assistant]   <tool_call>{"name": "image_search",
                 "arguments": {"url": "img_1"}}</tool_call>
[observation] [{'title': 'wedge-leaf hop-bush (Logan RE 12.8.20 Flora) -
                iNaturalist', 'source': 'iNaturalist'},
                {'title': 'AALII ...', 'source': '...'}, ...]
[assistant]   <tool_call>{"name": "text_search",
                 "arguments": {"q": "wedge-leaf hop-bush plant name"}}</tool_call>
[observation] [Passage 1] Title: Dodonaea viscosa  URL: https://en.wikipedia.org/
                wiki/Dodonaea_viscosa  Summary: Dodonaea viscosa, also known
                as the broadleaf hopbush ...
[assistant]   The plant with the pink seed pods is the wedge-leaf hop-bush,
              also known as Dodonaea viscosa ...
[stop]        No <tool_call> and no <response> found -- ending loop.
Final response: (empty)
GPT-5 judge:  correct=0, error_mode=no_answer.
\end{lstlisting}

\paragraph{Qwen3-VL-2B SFT.} The same backbone after one epoch of
full-trajectory SFT runs the same two-step search chain, but now closes
the rollout with a properly formatted \texttt{<response>}.
\begin{lstlisting}[style=prompt]
[user]        <image>What is the name of this plant with pink seed pods?
[assistant]   <think>Pink winged seed capsules on a green-leaf branch;
                image_search should identify the species.</think>
              <tool_call>{"name": "image_search",
                 "arguments": {"url": "img_1"}}</tool_call>
[observation] [{'title': 'wedge-leaf hop-bush (Logan RE 12.8.20 Flora)',
                'source': 'iNaturalist'}, ...]
[assistant]   <think>Hop-bush is the working hypothesis; confirm the
                scientific name with text_search.</think>
              <tool_call>{"name": "text_search",
                 "arguments": {"q": "wedge-leaf hop-bush scientific name"}}</tool_call>
[observation] [Passage 1] Title: Dodonaea viscosa  URL: https://en.wikipedia.org/
                wiki/Dodonaea_viscosa ...
[assistant]   <response>The plant with the pink seed pods is Dodonaea
              viscosa, commonly known as the Wedge-Leaf Hop Bush or
              Broadleaf Hop Bush ...</response>
GPT-5 judge:  correct=1, error_mode=none.
\end{lstlisting}

\paragraph{Interpretation.} Both rollouts retrieve the same evidence;
only the SFT model has internalized the contract that a final
\texttt{<response>} block is mandatory. This is the dominant
qualitative effect of full-trajectory SFT on 2B and the source of the
jump from 0.1\% to 28.4\% macro Pass@1 reported in Section~\ref{sec:results}.

%--------------------------------------------------------------------------
\subsection{Case 2: DPO recovers tool use on 2B (2B SFT $\to$ 2B DPO)}
\label{app:case-2b-dpo}

\begin{figure}[h]
\centering
\includegraphics[width=0.24\linewidth]{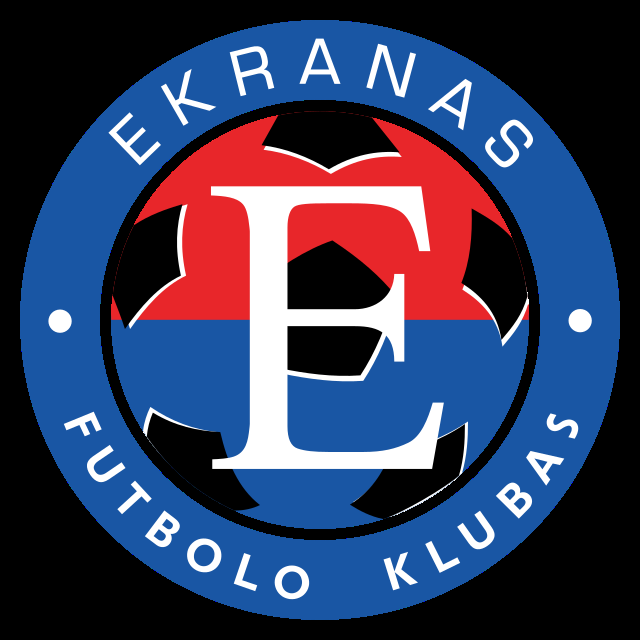}
\caption{Image for Case~\ref{app:case-2b-dpo} (FVQA
\texttt{fvqa\_004100}).}
\label{fig:case2-img}
\end{figure}

\paragraph{Setup.} Benchmark: FVQA, item \texttt{fvqa\_004100}. The
image is a circular football-club crest with the words
``\textsc{Ekranas}'' and ``\textsc{Futbolo Klubas}'' around a stylized
white ``E'' (Figure~\ref{fig:case2-img}).

\paragraph{Question.} ``What is the name of the football club shown
in the image?''

\paragraph{Reference.} \emph{FK Ekranas, a Lithuanian football club
(Futbolo Klubas Ekranas).}

\paragraph{Qwen3-VL-2B SFT.} The 2B SFT model skips tool use
entirely. It commits to an answer from raw perception alone,
misreading the Lithuanian word ``Futbolo Klubas'' as Bulgarian/Cyrillic
and inventing a fictitious Bulgarian club. The judge marks the answer
\emph{partial} because the entity name is right but the geographic
attribution is wrong.
\begin{lstlisting}[style=prompt]
[user]        <image>What is the name of the football club shown in the image?
[assistant]   <think>Circular crest with text I read as Cyrillic
                "EKranas / Futbolo Klubas".</think>
              <response>The football club shown in the image is Ekranas
              Futbolo Klubas (Bulgarian football club, based in Plovdiv) ...
              </response>
GPT-5 judge:  correct=0, partial=1, error_mode=partial -- correct entity,
              wrong country / city.
\end{lstlisting}

\paragraph{Qwen3-VL-2B DPO.} The DPO checkpoint invokes the tool
chain that the SFT model skipped: it identifies the crest with
\toolname{image\_search}, then issues a \toolname{text\_search} for
confirmation, and emits a correct \texttt{<response>} attributing the
club to Lithuania.
\begin{lstlisting}[style=prompt]
[user]        <image>What is the name of the football club shown in the image?
[assistant]   <think>[...]</think>
              <tool_call>{"name": "image_search",
                 "arguments": {"url": "img_1"}}</tool_call>
[observation] [{'title': 'FK Ekranas - Wikipedia', 'source': 'Wikipedia'},
                {'title': 'FK Ekranas 2025-26 Squad - ESPN (IN)',
                 'source': 'ESPN'}, ...]
[assistant]   <tool_call>{"name": "text_search",
                 "arguments": {"q": "FK Ekranas football club"}}</tool_call>
[observation] [Passage 1] Title: FK Ekranas  URL: https://en.wikipedia.org/wiki/
                FK_Ekranas  Summary: Ekranas is a Lithuanian football club ...
[assistant]   <response>The football club shown in the image is FK Ekranas
              (Futbolo Klubas Ekranas), a Lithuanian football club based in
              Panevezys ...</response>
GPT-5 judge:  correct=1, error_mode=none.
\end{lstlisting}

\paragraph{Interpretation.} The 2B SFT model exhibits the
\emph{premature answer} failure mode that the synthetic hard-negative
classes target directly (Section~\ref{sec:method-dpo}): it commits
from raw perception without invoking any tool. DPO shifts the same
backbone toward the \emph{evidence-then-answer} contract.

%--------------------------------------------------------------------------
\subsection{Case 3: DPO closes the loop on a multi-tool rollout
(4B SFT $\to$ 4B DPO)}
\label{app:case-4b-dpo-closure}

\paragraph{Setup.} Benchmark: SimpleVQA, item \texttt{simplevqa\_1057}.
The image is a Costco retail wine display showing three shelf price
tags: \textsc{Northstar Merlot} (Columbia Valley, 3.0\,L) at \$139.99,
\textsc{Mo\"et \& Chandon Imp\'erial} (1.5\,L) at \$114.99, and
\textsc{Chateau Ste.\ Michelle Cold Creek Cabernet} (Columbia Valley,
3.0\,L) at \$109.99. The question asks only about the Merlot and
Cabernet tags (Figure~\ref{fig:case3-img}).

\begin{figure}[h]
\centering
\includegraphics[width=0.32\linewidth]{cases/case3_simplevqa_wines.jpg}
\caption{Image for Case~\ref{app:case-4b-dpo-closure} (SimpleVQA
\texttt{simplevqa\_1057}).}
\label{fig:case3-img}
\end{figure}

\paragraph{Question.} ``What is the total price for a bottle of Merlot
and a bottle of Cabernet shown in the image?''

\paragraph{Reference.} \emph{\$249.98.}

\paragraph{Qwen3-VL-4B SFT.} The SFT model chains several
image-processing and search tools---\toolname{crop},
\toolname{super\_resolution}, and finally \toolname{text\_search}---never
produces a final answer, and exhausts the step budget.
\begin{lstlisting}[style=prompt]
[user]        <image>What is the total price for a bottle of Merlot and a
                bottle of Cabernet shown in the image?
[...]
[assistant]   <think>Crop the top of the image to isolate the price tags.</think>
              <tool_call>{"name": "crop",
                 "arguments": {"image": "img_1", "x": 0, "y": 0,
                               "width": 300, "height": 200}}</tool_call>
[observation] Image cropped successfully. New image ID: img_2. ...
[assistant]   <think>[...]</think>
              <tool_call>{"name": "super_resolution",
                 "arguments": {"image": "img_2", "scale": 4}}</tool_call>
[observation] Super resolution enhancement completed. New image ID: img_3. ...
[...]
[assistant]   <think>[...]</think>
              <tool_call>{"name": "text_search",
                 "arguments": {"q": "Costco Merlot Cabernet price"}}</tool_call>
[observation] [Passage 1] Title: A $13.99 Costco Wine Find ...
[stop]        Hit max_steps=5 without a <response>.
GPT-5 judge:  correct=0, error_mode=no_answer.
\end{lstlisting}

\paragraph{Qwen3-VL-4B DPO.} The DPO checkpoint crops the price-tag
region, re-reads the two visible prices from the cropped image
(\$139.99 and \$109.99), runs a sanity-check \toolname{text\_search},
and commits to a correct final answer with arithmetic.
\begin{lstlisting}[style=prompt]
[user]        <image>What is the total price for a bottle of Merlot and a
                bottle of Cabernet shown in the image?
[assistant]   <think>Crop the top-left price tag region to read the prices
                directly.</think>
              <tool_call>{"name": "crop",
                 "arguments": {"image": "img_1", "x": 0, "y": 0,
                               "width": 300, "height": 200}}</tool_call>
[observation] Image cropped successfully. New image ID: img_2. ...
[assistant]   <think>From img_2 I can read: MERLOT $139.99, CABERNET
                $109.99. Sanity-check with text_search before committing.
                </think>
              <tool_call>{"name": "text_search",
                 "arguments": {"q": "Costco Merlot Cabernet store price"}}</tool_call>
[observation] [Passage 1] Title: Total Wine & More ...
[assistant]   <response>Based on the image, the prices are Merlot $139.99
              and Cabernet $109.99. The total is $139.99 + $109.99 = $249.98.
              \boxed{249.98}</response>
GPT-5 judge:  correct=1, error_mode=none.
\end{lstlisting}

\paragraph{Interpretation.} The two failure modes the SFT model
exhibits here---\emph{wrong tool} (chaining \toolname{super\_resolution}
on top of \toolname{crop} instead of reading the visible prices) and
\emph{repeated query} (re-asking the same visual question without
committing to the visible numbers)---are exactly the categories
targeted by the synthetic step-DPO negatives (Section~\ref{sec:method-dpo}).
The DPO checkpoint commits earlier and uses the visible image
evidence: it crops once, reads the prices, and writes the final
answer. This is the multi-tool source of the SimpleVQA improvements
visible in Table~\ref{tab:main-pass1}.

%--------------------------------------------------------------------------
\subsection{Case 4: DPO avoids over-specification (4B SFT-partial $\to$ 4B DPO-correct)}
\label{app:case-4b-dpo-precision}

\begin{figure}[h]
\centering
\includegraphics[width=0.42\linewidth]{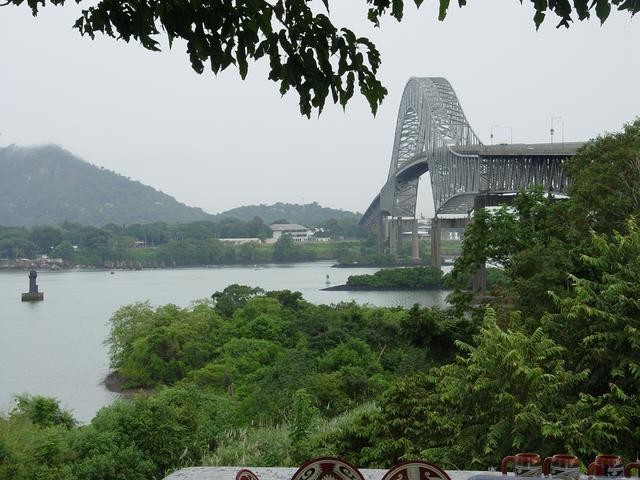}
\caption{Image for Case~\ref{app:case-4b-dpo-precision} (FVQA
\texttt{fvqa\_004018}).}
\label{fig:case4-img}
\end{figure}

\paragraph{Setup.} Benchmark: FVQA, item \texttt{fvqa\_004018}. The
image shows a steel-arch road bridge spanning a tropical waterway,
with green hills in the background (Figure~\ref{fig:case4-img}).

\paragraph{Question.} ``In which year did this bridge officially
open?''

\paragraph{Reference.} \emph{1962 (the Bridge of the Americas / Puente
de las Am\'ericas, Panama, officially opened on October 12, 1962).}

\paragraph{Qwen3-VL-4B SFT.} The model retrieves correct evidence
identifying the Bridge of the Americas in Panama and the opening year
1962. It then \emph{over-specifies}: instead of stopping at the
requested year, it inserts a fabricated exact date (``October 2,
1962'') taken from a partial reading of the search passage. The judge
marks the answer as \emph{partial} because the year is right but the
day is wrong by ten days.
\begin{lstlisting}[style=prompt]
[user]        <image>In which year did this bridge officially open?
[assistant]   <think>Steel-arch road bridge over a tropical channel; identify
                via image_search.</think>
              <tool_call>{"name": "image_search",
                 "arguments": {"url": "img_1"}}</tool_call>
[observation] [{'title': 'Ficheiro:PuenteAmericas3.jpg - Wikipedia, a
                enciclopedia libre', 'source': 'Wikipedia'},
                {'title': 'The Bridge of the Americas is a road bridge
                in Panama ...', 'source': 'Alamy'}, ...]
[assistant]   <tool_call>{"name": "text_search",
                 "arguments": {"q": "Bridge of the Americas Panama opening
                                year"}}</tool_call>
[observation] [Passage 1] Title: Bridge of the Americas
                URL: https://en.wikipedia.org/wiki/Bridge_of_the_Americas
                Summary: ... officially opened on October 12, 1962 ...
[assistant]   <response>This is the Bridge of the Americas (Puente de las
              Americas) in Panama. According to historical records, this
              bridge officially opened on October 2, 1962 ...</response>
GPT-5 judge:  correct=0, partial=1, error_mode=partial -- correct bridge
              and year, but wrong specific day.
\end{lstlisting}

\paragraph{Qwen3-VL-4B DPO.} The DPO checkpoint executes the same
two tool calls but answers at the granularity the question actually
requests, dropping the unrequested day-level detail.
\begin{lstlisting}[style=prompt]
[user]        <image>In which year did this bridge officially open?
[assistant]   <tool_call>{"name": "image_search",
                 "arguments": {"url": "img_1"}}</tool_call>
[observation] [{'title': 'The Bridge of the Americas is a road bridge in
                Panama ...', 'source': 'Alamy'}, ...]
[assistant]   <tool_call>{"name": "text_search",
                 "arguments": {"q": "Bridge of the Americas Panama opening
                                year"}}</tool_call>
[observation] [Passage 1] Title: Bridge of the Americas  Summary:
                Completed in 1962 at a cost of US$20 million ...
[assistant]   <response>This is the Bridge of the Americas (Puente de las
              Americas) in Panama. It officially opened in 1962 ...
              </response>
GPT-5 judge:  correct=1, error_mode=none.
\end{lstlisting}

\paragraph{Interpretation.} The SFT model adds a specific date that
the question does not ask for and that does not survive scrutiny. The
DPO checkpoint terminates at the granularity supported by the
question and the retrieved evidence. This is the same answer-fidelity
mechanism the synthetic step-DPO negatives target through the
\emph{premature answer} and \emph{ignored image} classes
(Section~\ref{sec:method-dpo}): the pre-DPO model is willing to
elaborate beyond what evidence justifies, while the post-DPO model
stays inside the supported envelope. This pattern contributes to the
+1.7 average-point gain at 4B (Section~\ref{sec:results}).

%--------------------------------------------------------------------------
\subsection{Case 5: A failure that survives DPO --- verification on VDR}
\label{app:case-vdr-failure}

\begin{figure}[h]
\centering
\includegraphics[width=0.24\linewidth]{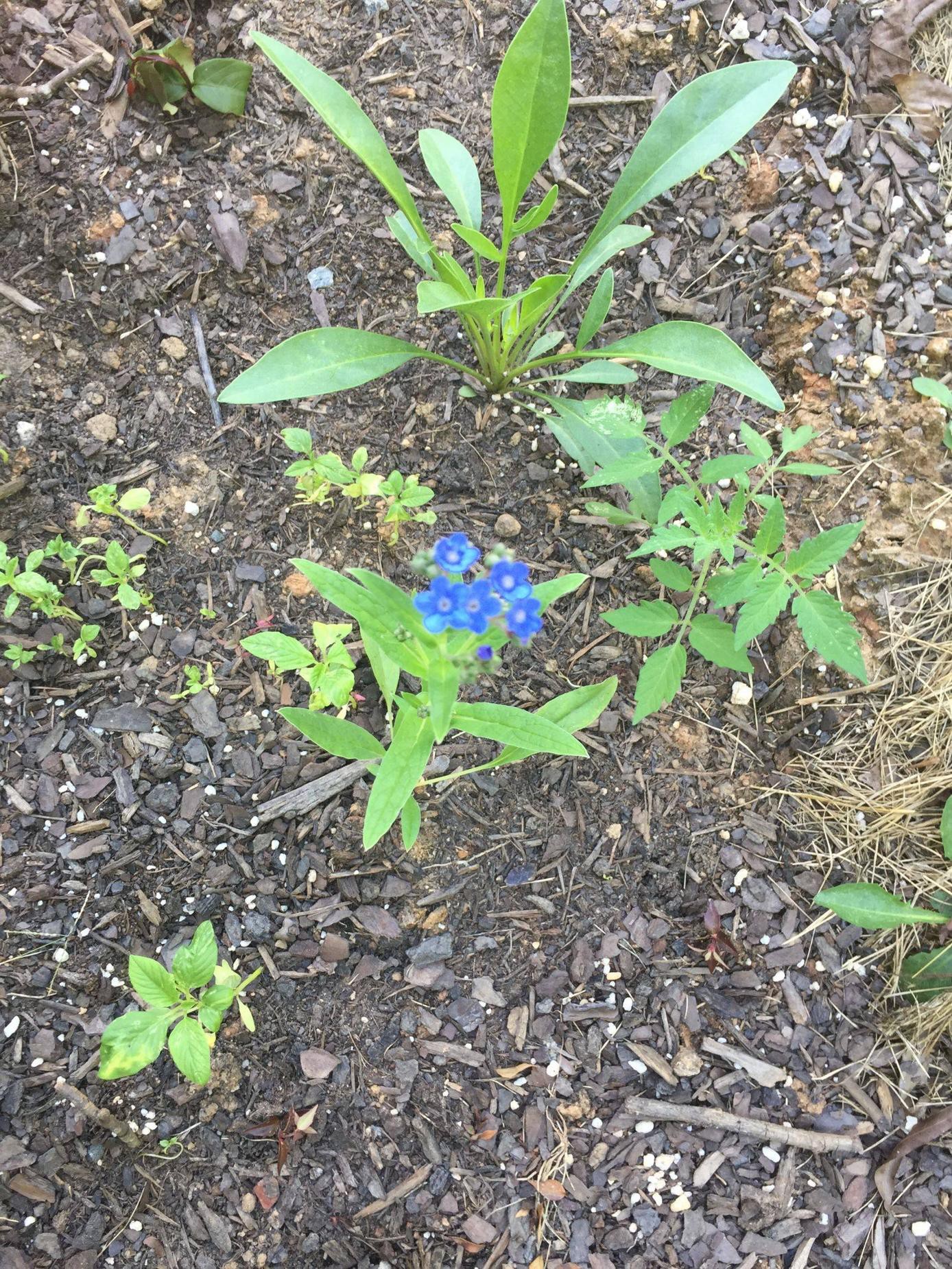}
\caption{Image for Case~\ref{app:case-vdr-failure} (VDR
\texttt{vdr\_base\_Nature\_833}).}
\label{fig:case5-img}
\end{figure}

\paragraph{Setup.} Benchmark: VDR, item
\texttt{vdr\_base\_Nature\_833}. The image is an overhead photo of a
garden bed; at the top is a leafy plant forming a rosette of broad,
spear-shaped leaves, with a small blue forget-me-not flower and
seedlings nearby (Figure~\ref{fig:case5-img}).

\paragraph{Question.} ``What is the name of the plant at the top with
broad, spear-shaped leaves?''

\paragraph{Reference.} \emph{Coreopsis.}

\paragraph{Qwen3-VL-4B SFT.} The SFT model crops the leaf rosette,
then guesses a different species in each search
(\emph{forget-me-not}, then \emph{columbine}) without ever committing
to a final answer, and exhausts the step budget.
\begin{lstlisting}[style=prompt]
[user]        <image>What is the name of the plant at the top with broad,
                spear-shaped leaves?
[...]
[assistant]   <think>Crop the top rosette to inspect the leaves.</think>
              <tool_call>crop(img_1, x=100, y=0, width=300, height=200)</tool_call>
[observation] Image cropped successfully. New image ID: img_2. ...
[...]
[assistant]   <tool_call>text_search("plant spear-shaped leaves garden bed
                forget-me-not companion plant")</tool_call>
[observation] ...
[assistant]   <tool_call>text_search("Columbine plant spear-shaped leaves
                garden bed")</tool_call>
[observation] ...
[stop]        Hit max_steps=5 without a <response>.
GPT-5 judge:  correct=0, error_mode=no_answer.
\end{lstlisting}

\paragraph{Qwen3-VL-4B DPO.} The DPO checkpoint commits early, but to a confident misidentification: it names \emph{Broadleaf
Plantain} (\emph{Plantago major}) --- a plausible visual match for a
broad-leaved rosette, but the wrong species.
\begin{lstlisting}[style=prompt]
[user]        <image>What is the name of the plant at the top with broad,
                spear-shaped leaves?
[...]
[assistant]   <tool_call>text_search("garden plant broad spear-shaped leaves
                dandelion plantain identification")</tool_call>
[observation] ...
[assistant]   <response>The plant at the top with broad, spear-shaped leaves
              is Broadleaf Plantain (Plantago major), also commonly known as
              Common Plantain ...</response>
GPT-5 judge:  correct=0, error_mode=wrong_entity.
\end{lstlisting}

\paragraph{Interpretation.} DPO has done its job in one sense
(\texttt{no\_answer} $\to$ a committed answer), but the answer is a
plausible-but-wrong species: the DPO model seeds its search
with a guessed identity (``dandelion plantain'') and commits to the
first plausible match without cross-checking the leaf morphology
against the true species. This is the canonical instance of the
verification bottleneck identified in Section~\ref{sec:results-vdr}:
on multi-hop deep-research questions, additional search turns reliably
convert \texttt{no\_answer} into \texttt{wrong\_entity} rather than
into \texttt{correct}. Closing this gap requires a verification-style
signal that the current SFT and synthetic-DPO recipes do not provide.

\section{Broader Impacts}
\label{app:impacts}
Making tool-using multimodal search agents workable at 2B--4B parameters lowers the compute and cost barrier to deploying them, which is broadly useful for on-device and low-resource settings and is the motivation for this work. The same accessibility carries a risk that our own error analysis makes concrete: the resulting agents retrieve real web evidence yet still commit confidently to a wrong primary entity on a substantial fraction of questions (Section~4, Appendix~K), and the VDR ablation shows that giving them more search turns converts abstentions into confident errors rather than correct answers. Deployments should therefore not treat these agents' answers as verified, and we identify verification-aware training as the priority for future work. We release no new model weights or scraped corpora with this submission.

%--------------------------------------------------------------------------
% NOTE: For a NeurIPS main-track submission, the NeurIPS paper checklist
% would be inserted here (e.g., \input{checklist}). The workshop CFP does
% not require the checklist, so it is intentionally omitted.
%--------------------------------------------------------------------------

\end{document}